\pdfoutput=1
\documentclass{article}

\usepackage[final]{nips_2017}

\usepackage[utf8]{inputenc} 
\usepackage[T1]{fontenc}    
\usepackage{hyperref}       
\usepackage{url}            
\usepackage{booktabs}       
\usepackage{amsfonts}       
\usepackage{nicefrac}       
\usepackage{microtype}      
\usepackage{xcolor}
\usepackage{hyperref}
\usepackage{graphicx}
\usepackage{algorithmicx}
\usepackage{algorithm}
\usepackage{algpseudocode}
\usepackage{mathrsfs}
\usepackage{amssymb}
\usepackage{graphicx,wrapfig,lipsum}
\usepackage{natbib}
\usepackage{caption}
\usepackage{subcaption}
\usepackage{listings}
\usepackage{cleveref}

\newif\ifshort
\shorttrue

\ifshort
    \usepackage{titlesec}
    \titlespacing\section{0pt}{0pt plus 4pt minus 2pt}{0pt plus 2pt minus 2pt}
    \titlespacing\subsection{0pt}{0pt plus 4pt minus 2pt}{0pt plus 2pt minus 2pt}
\else
    
\fi

\usepackage{lipsum}

\title{Multi-Goal Reinforcement Learning: Challenging Robotics Environments and Request for Research}

\author{
    Matthias Plappert, Marcin Andrychowicz, Alex Ray, Bob McGrew,\\ \textbf{Bowen Baker, Glenn Powell, Jonas Schneider, Josh Tobin,}\\
    \textbf{Maciek Chociej, Peter Welinder, Vikash Kumar, and Wojciech Zaremba}\\
  OpenAI \\ ~ \\
  Correspondence to \texttt{\{matthias, marcin\}@openai.com} \\
}

\begin{document}
\maketitle

\begin{abstract}

The purpose of this technical report is two-fold.
First of all,
it introduces a suite of challenging continuous control tasks (integrated with OpenAI~Gym)
based on currently existing robotics hardware.
The tasks include pushing, sliding and pick \& place with a Fetch
robotic arm as well as in-hand object manipulation with a
Shadow Dexterous Hand.
All tasks have sparse binary rewards and follow a Multi-Goal Reinforcement Learning (RL) framework in which an agent
is told what to do using an additional input.

The second part of the paper presents a set of concrete research ideas for improving RL algorithms,
most of which are related to Multi-Goal RL and Hindsight Experience Replay.

\end{abstract}

\section{Environments}

All environments are released as part of \emph{OpenAI Gym}\footnote{\url{https://github.com/openai/gym}} \citep{gym}
and use the \emph{MuJoCo} \citep{MuJoCo} physics engine for fast and accurate simulation.
A video presenting the new environments can be found at \url{https://www.youtube.com/watch?v=8Np3eC_PTFo}.

\subsection{Fetch environments}

The Fetch environments are based on
the $7$-DoF Fetch robotics arm,\footnote{\url{http://fetchrobotics.com/}}
which has a two-fingered parallel gripper.
They are very similar to the tasks used in \cite{her} but we have added an additional \emph{reaching} task and the \emph{pick \& place} task
is a bit different.\footnote{In \cite{her}
training on this task relied on starting some of the training
episodes from a state in which the box is already grasped.
This is not necessary for successful training if the
target position of the box is sometimes in the air and sometimes on the
table and we do not use this technique anymore.}

In all Fetch tasks, the goal is $3$-dimensional and describes the desired position of the object (or the end-effector for reaching). Rewards are sparse and binary: The agent obtains a reward of $0$ if the object is at the target location (within a tolerance of $5$~cm) and $-1$ otherwise.
Actions are $4$-dimensional: $3$ dimensions specify the desired gripper movement in Cartesian coordinates and the last dimension controls opening and closing of the gripper.
We apply the same action in $20$ subsequent simulator steps (with $\Delta t = 0.002$ each) before returning control to the agent, i.e. the agent's action frequency is $f = 25$~Hz.
Observations include the Cartesian position of the gripper, its linear velocity as well as the position and linear velocity of the robot's gripper.
If an object is present, we also include the object's Cartesian position and rotation using Euler angles, its linear and angular velocities, as well as its position and linear velocities relative to gripper.

\begin{figure}
    \centering
    \begin{subfigure}[b]{0.23\textwidth}
        \includegraphics[width=\textwidth]{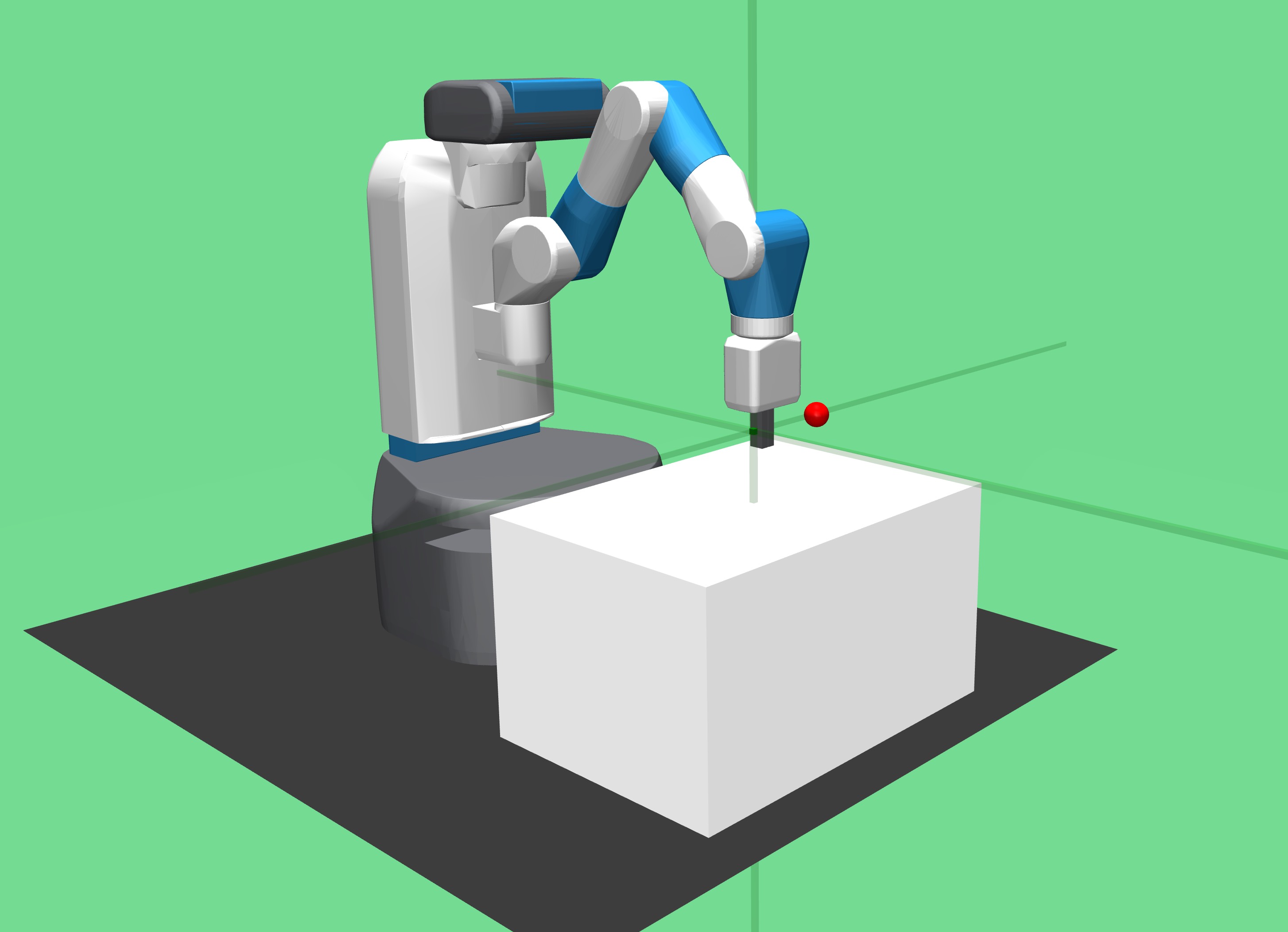}
    \end{subfigure}
    ~
    \begin{subfigure}[b]{0.23\textwidth}
        \includegraphics[width=\textwidth]{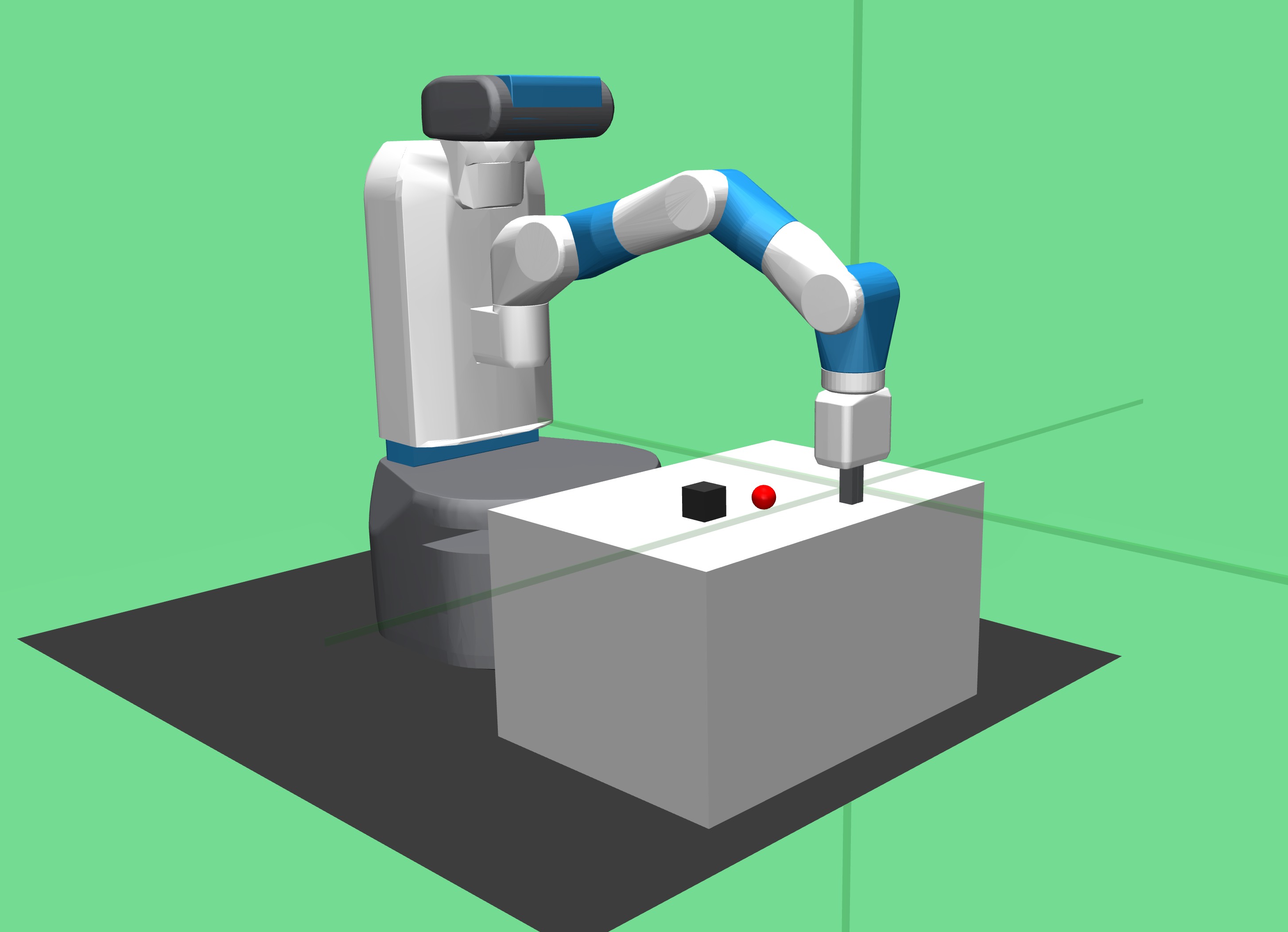}
    \end{subfigure}
    ~
    \begin{subfigure}[b]{0.23\textwidth}
        \includegraphics[width=\textwidth]{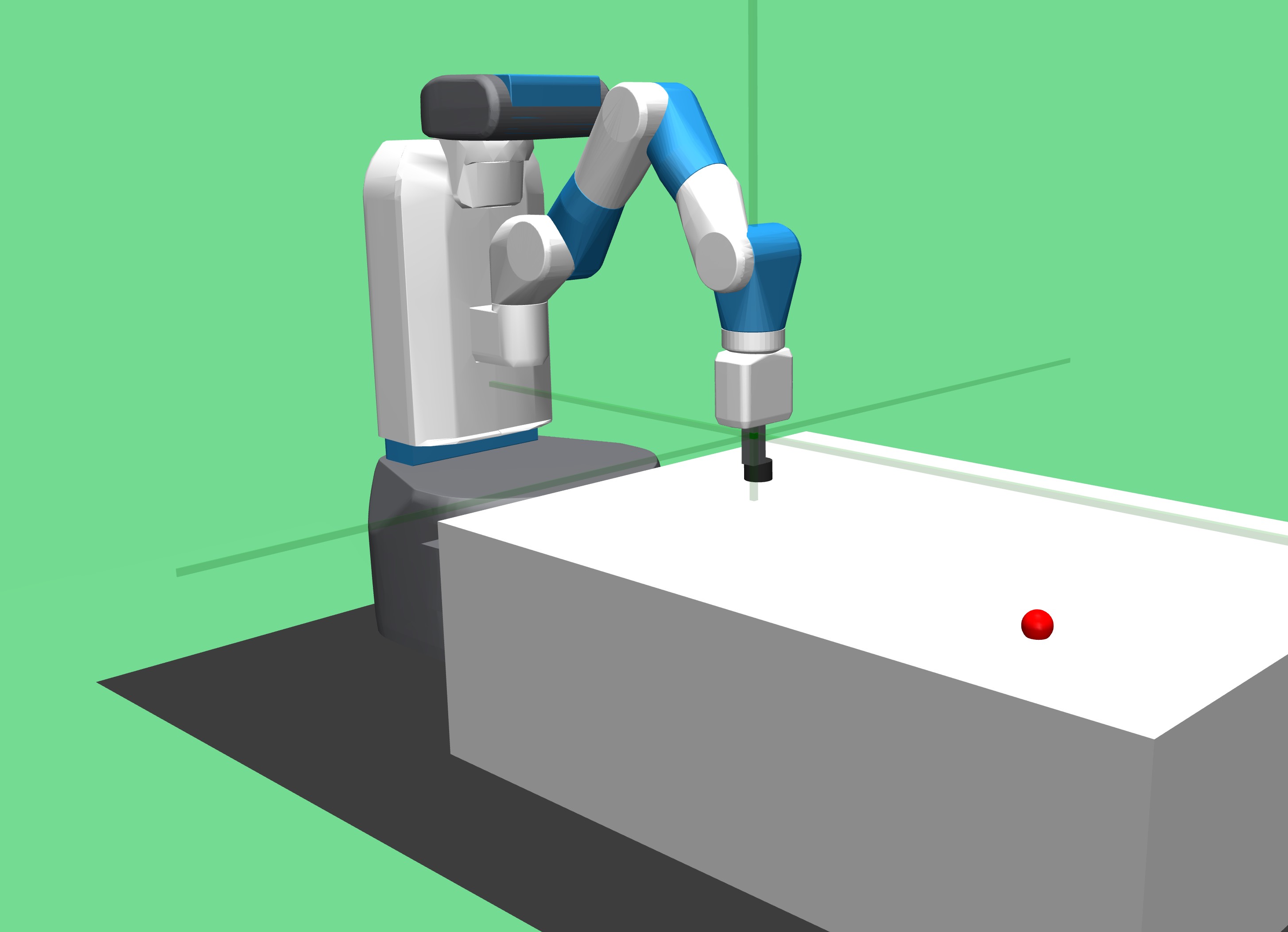}
    \end{subfigure}
    ~
    \begin{subfigure}[b]{0.23\textwidth}
        \includegraphics[width=\textwidth]{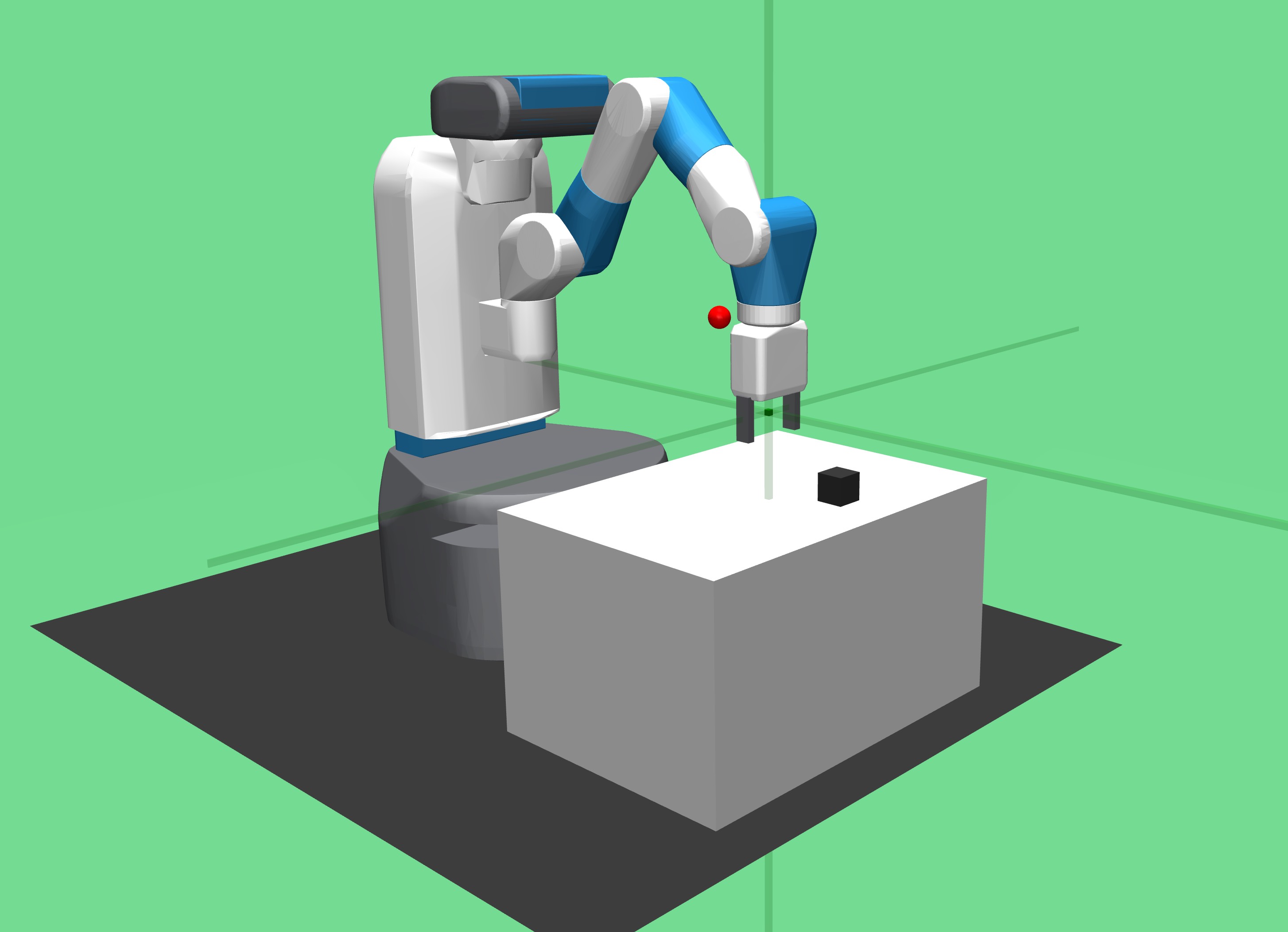}
    \end{subfigure}
    \caption{The four proposed Fetch environments: \texttt{FetchReach}, \texttt{FetchPush}, \texttt{FetchSlide}, and \texttt{FetchPickAndPlace}.}
    \label{fig:hand-envs}
\end{figure}

\paragraph{Reaching (\texttt{FetchReach})}
The task is to move the gripper to a target position.
This task is very easy to learn and is therefore a suitable benchmark to ensure that a new idea works at all.\footnote{That being said, we have found that is so easy that even partially broken implementations sometimes learn successful policies, so no conclusions should be drawn from this task alone.}

\paragraph{Pushing (\texttt{FetchPush})}
A box is placed on a table in front of the robot and the task is to move it to a target location on the table.
The robot fingers are locked to prevent grasping. The learned behavior is usually a mixture of pushing and rolling.

\paragraph{Sliding (\texttt{FetchSlide})} A puck is placed on a long slippery table
and the target position is outside of the robot's reach so that it has
to hit the puck with such a force that it slides and then stops at the target location due to friction.

\paragraph{Pick \& Place (\texttt{FetchPickAndPlace})} The task is to grasp
a box and move it to the target location which
may be located on the table surface or in the air above it.

\subsection{Hand environments}

These environments are based on the Shadow Dexterous Hand,\footnote{https://www.shadowrobot.com/products/dexterous-hand/} which is an anthropomorphic robotic hand with 24 degrees of freedom.
Of those 24 joints, 20 can be can be controlled independently whereas the remaining ones are coupled joints.

In all hand tasks, rewards are sparse and binary: The agent obtains a reward of $0$ if the goal has been achieved (within some task-specific tolerance) and $-1$ otherwise.
Actions are $20$-dimensional: We use absolute position control for all non-coupled joints of the hand.
We apply the same action in $20$ subsequent simulator steps (with $\Delta t = 0.002$ each) before returning control to the agent, i.e. the agent's action frequency is $f = 25$~Hz.
Observations include the $24$ positions and velocities of the robot's joints.
In case of an object that is being manipulated, we also include its Cartesian position and rotation represented by a quaternion (hence $7$-dimensional) as well as its linear and angular velocities.
In the reaching task, we include the Cartesian position of all $5$ fingertips.

\paragraph{Reaching (\texttt{HandReach})}
A simple task in which the goal is $15$-dimensional and contains the target Cartesian position of each fingertip of the hand.
Similarly to the \texttt{FetchReach} task, this task is relatively easy to learn.
A goal is considered achieved if the mean distance between fingertips and their desired position is less than $1$~cm.

\paragraph{Block manipulation (\texttt{HandManipulateBlock})}
In the block manipulation task, a block is placed on the palm of the hand.
The task is to then manipulate the block such that a target pose is achieved.
The goal is $7$-dimensional and includes the target position (in Cartesian coordinates) and target rotation (in quaternions).
We include multiple variants with increasing levels of difficulty:
\begin{itemize}
    \item \texttt{HandManipulateBlockRotateZ} Random target rotation around the $z$ axis of the block. No target position.
    \item \texttt{HandManipulateBlockRotateParallel} Random target rotation around the $z$ axis of the block and axis-aligned target rotations for the $x$ and $y$ axes. No target position.
    \item \texttt{HandManipulateBlockRotateXYZ} Random target rotation for all axes of the block. No target position.
    \item \texttt{HandManipulateBlockFull} Random target rotation for all axes of the block. Random target position.
\end{itemize}
A goal is considered achieved if the distance between the block's position and its desired position is less than $1$~cm (applicable only in the \texttt{Full} variant) and the difference in rotation is less than $0.1$~rad.

\paragraph{Egg manipulation (\texttt{HandManipulateEgg})}
The objective here is similar to the block task but instead of a block an egg-shaped object is used.
We find that the object geometry makes a significant differences in how hard the problem is and the egg is probably the easiest object.
The goal is again $7$-dimensional and includes the target position (in Cartesian coordinates) and target rotation (in quaternions).
We include multiple variants with increasing levels of difficulty:
\begin{itemize}
    \item \texttt{HandManipulateEggRotate} Random target rotation for all axes of the egg. No target position.
    \item \texttt{HandManipulateEggFull} Random target rotation for all axes of the egg. Random target position.
\end{itemize}
A goal is considered achieved if the distance between the egg's position and its desired position is less than $1$~cm (applicable only in the \texttt{Full} variant) and the difference in rotation is less than $0.1$~rad.

\paragraph{Pen manipulation (\texttt{HandManipulatePen})}
Another manipulation, this time using a pen instead of a block or an egg.
Grasping the pen is quite hard since it easily falls off the hand and can easily collide and get stuck between other fingers.
The goal is $7$-dimensional and includes the target position (in Cartesian coordinates) and target rotation (in quaternions).
We include multiple variants with increasing levels of difficulty:
\begin{itemize}
    \item \texttt{HandManipulatePenRotate} Random target rotation $x$ and $y$ axes of the pen and no target rotation around the $z$ axis. No target position.
    \item \texttt{HandManipulatePenFull} Random target rotation $x$ and $y$ axes of the pen and no target rotation around the $z$ axis. Random target position.
\end{itemize}
A goal is considered achieved if the distance between the pen's position and its desired position is less than $5$~cm (applicable only in the \texttt{Full} variant) and the difference in rotation, ignoring the $z$ axis,\footnote{The $z$ axis of the pen is parallel to its body and goes through its tip to its opposite end.} is less than $0.1$~rad.

\begin{figure}
    \centering
    \begin{subfigure}[b]{0.23\textwidth}
        \includegraphics[width=\textwidth]{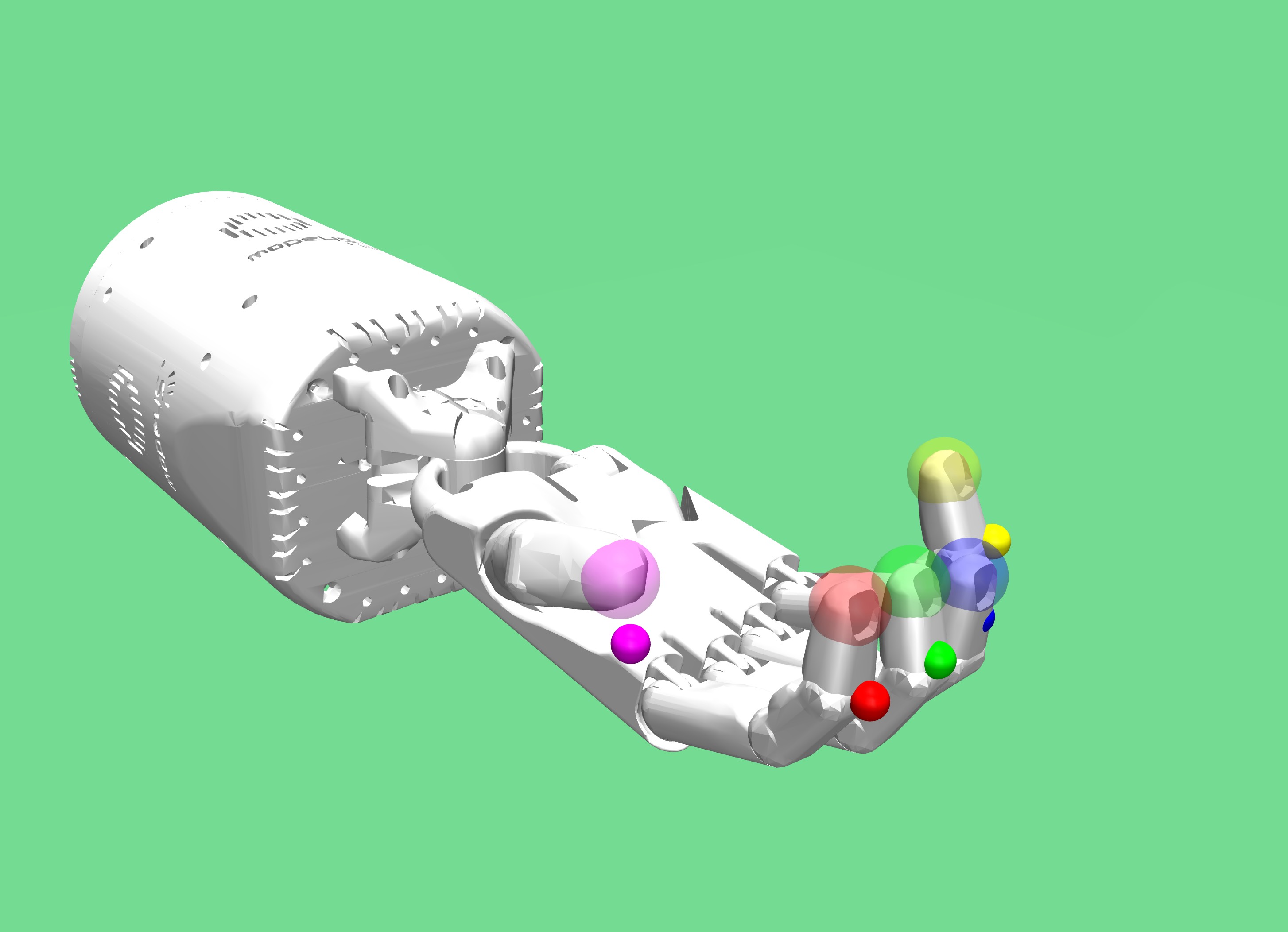}
    \end{subfigure}
    ~
    \begin{subfigure}[b]{0.23\textwidth}
        \includegraphics[width=\textwidth]{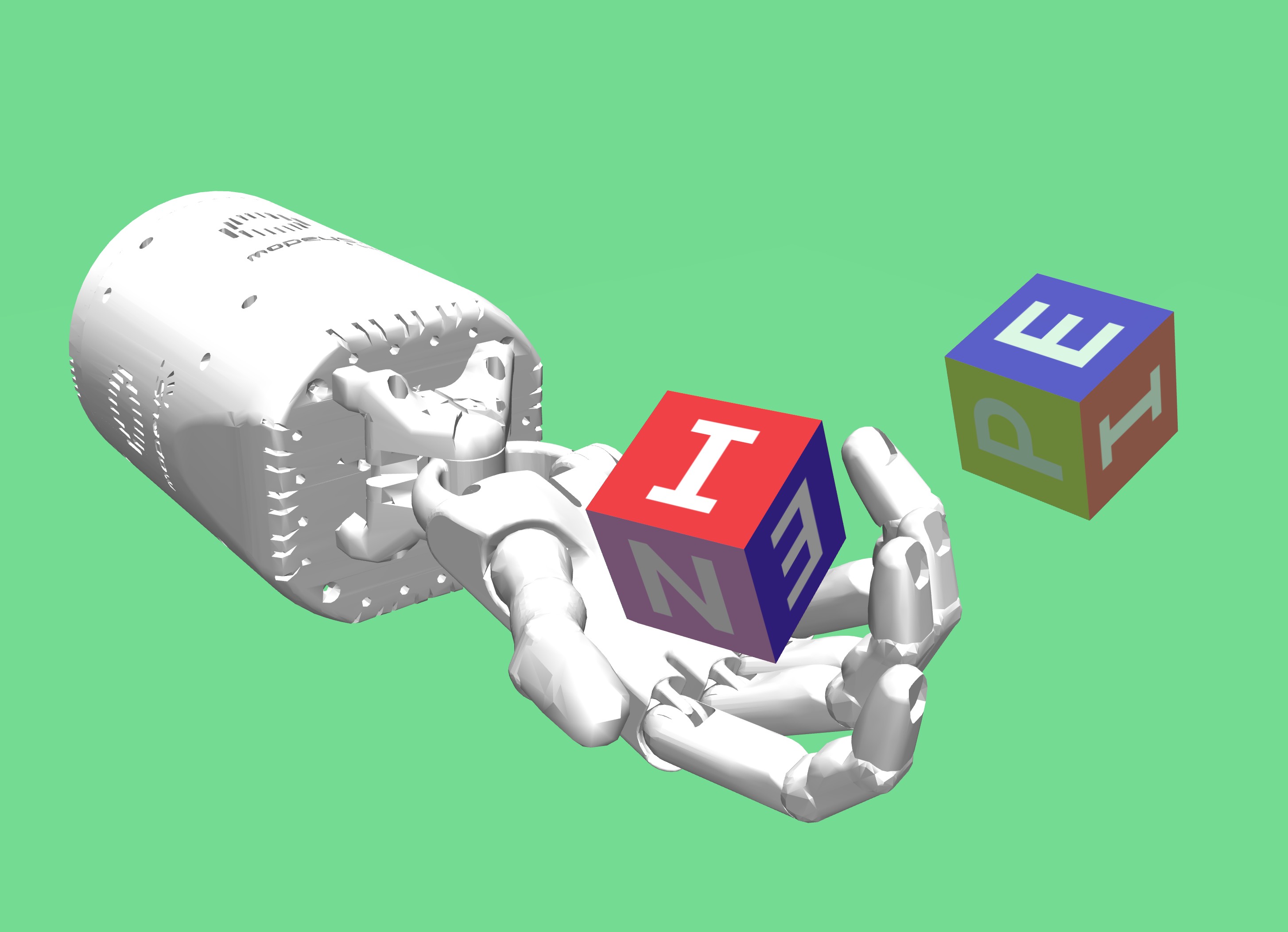}
    \end{subfigure}
    ~
    \begin{subfigure}[b]{0.23\textwidth}
        \includegraphics[width=\textwidth]{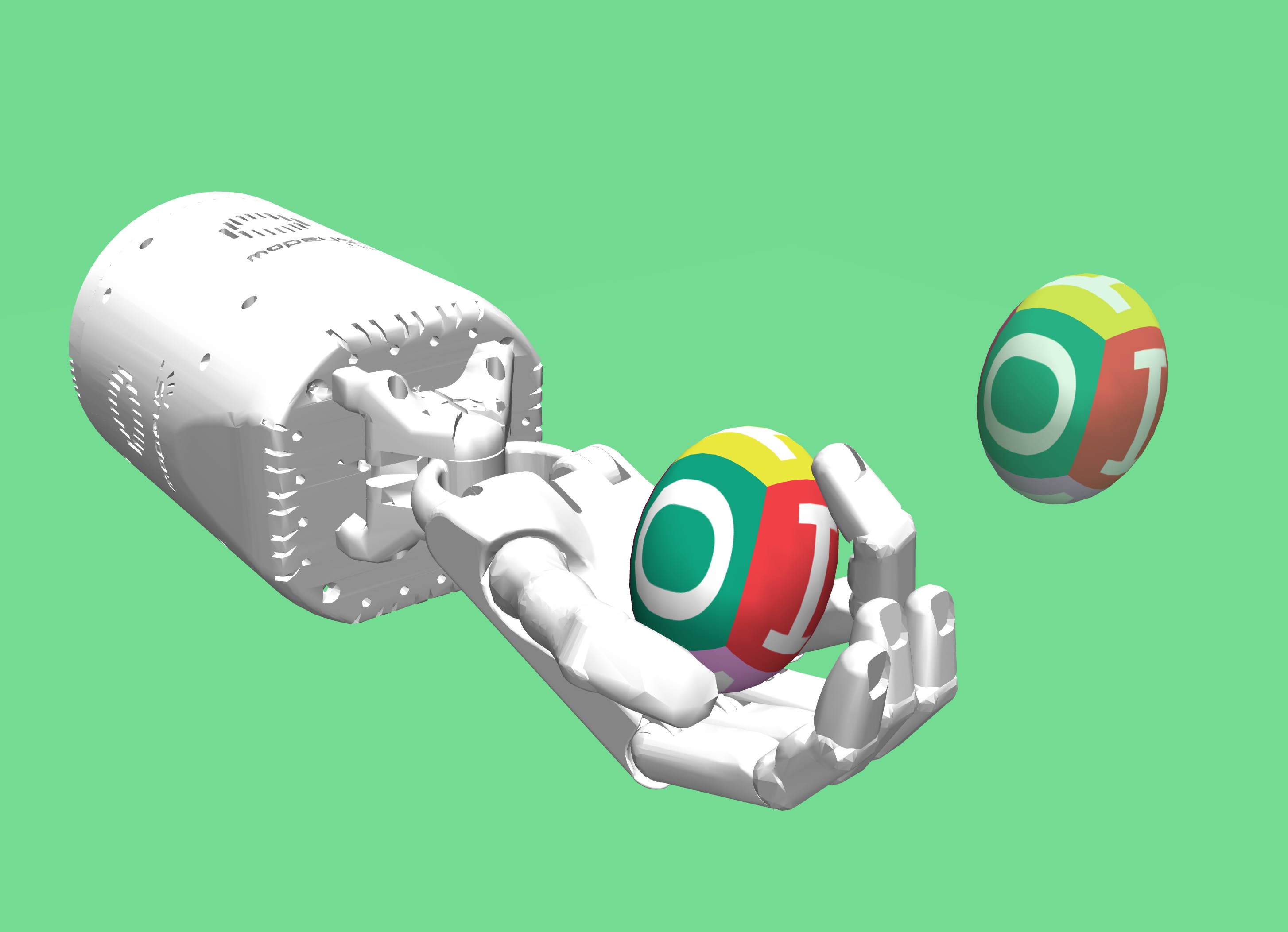}
    \end{subfigure}
    ~
    \begin{subfigure}[b]{0.23\textwidth}
        \includegraphics[width=\textwidth]{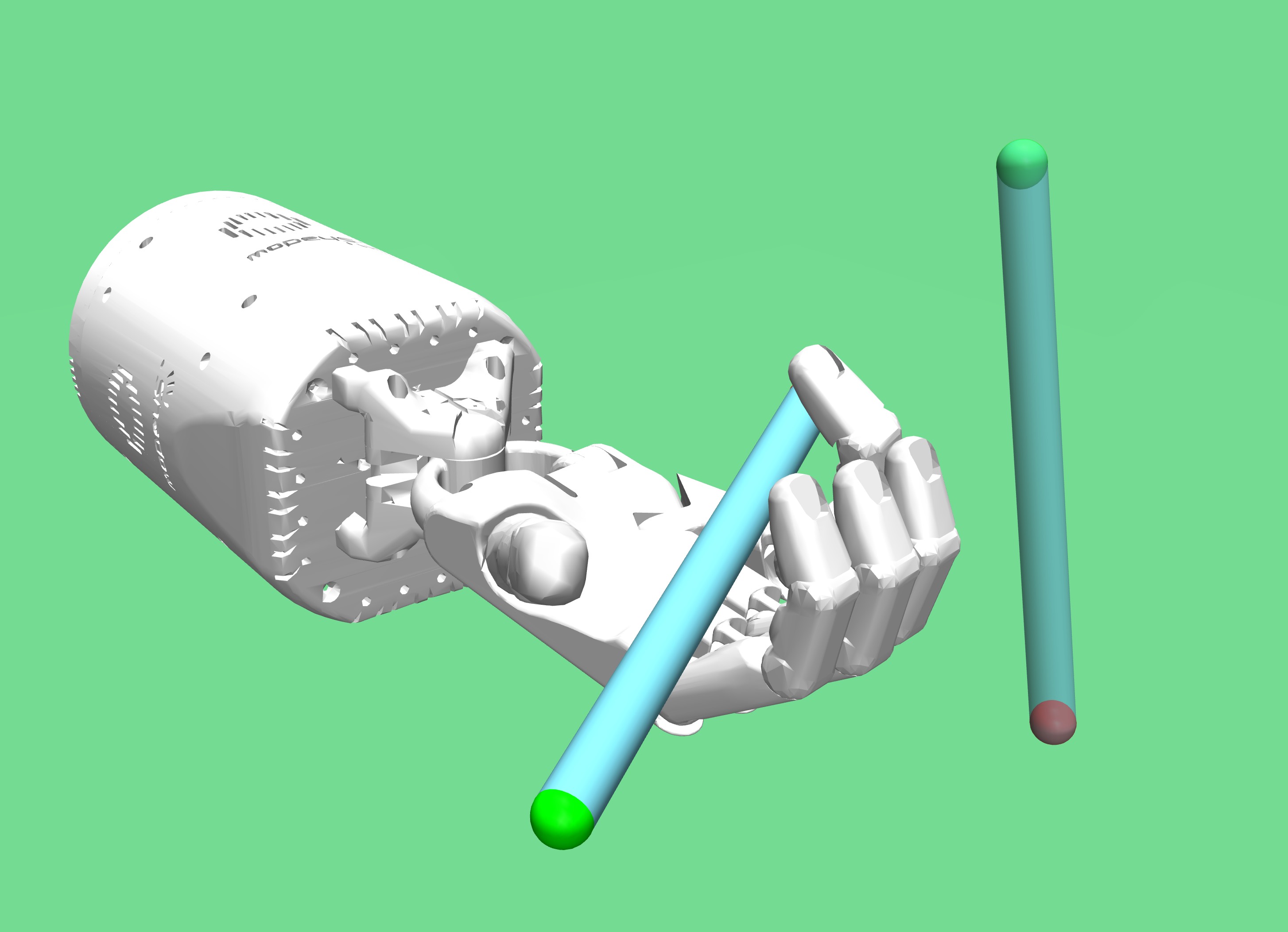}
    \end{subfigure}
    \caption{The four proposed Shadow Dexterous Hand environments: \texttt{HandReach}, \texttt{HandManipulateBlock}, \texttt{HandManipulateEgg}, and \texttt{HandManipulatePen}.}
    \label{fig:hand-envs}
\end{figure}

\subsection{Multi-goal environment interface}

All environments use \emph{goals} that describe the desired outcome of a task.
For example, in the \texttt{FetchReach} task, the desired target position is described by a $3$-dimensional goal.
While our environments are fully compatible with the OpenAI Gym API, we slightly extend upon it to support this new type of environment. All environments extend the newly introduced \texttt{gym.GoalEnv}.

\paragraph{Goal-aware observation space}

First, it enforces a constraint on the observation space.
More concretely, it requires that the observation space is of type \texttt{gym.spaces.Dict} space, with at least the following three keys:
\begin{itemize}
    \item \texttt{observation}: The actual observation of the environment, For example robot state and position of objects.
    \item \texttt{desired\_goal}: The goal that the agent has to achieve. In case of \texttt{FetchReach}, this would be the 3-dimensional target position.
    \item \texttt{achieved\_goal}: The goal that the agent has currently achieved instead. In case of \texttt{FetchReach}, this is the position of the robots end effector. Ideally, this would be the same as \texttt{desired\_goal} as quickly as possible.
\end{itemize}

\paragraph{Exposed reward function}

Second, we expose the reward function in a way that allows for re-computing the reward with different goals.
This is a necessary requirement for HER-style algorithms which substitute goals.
A detailed example is available in \Cref{appendix:code-examples}

\paragraph{Compatibility with standard RL algorithms}
Since OpenAI Gym is commonly supported in most RL algorithm frameworks and tools like OpenAI Baselines~\citep{baselines}, we include a simple wrapper that converts the new dictionary-based goal observation space into a more common array representation.
A detailed example is available in \Cref{appendix:code-examples}.

\subsection{Benchmark results}

We evaluate the performance of DDPG with and without Hindsight Experience Replay~(HER,~\cite{her}) on all environments with all its variants.
We compare the following four configurations:
\begin{itemize}
    \item DDPG+HER with sparse rewards
    \item DDPG+HER with dense rewards
    \item DDPG with sparse rewards
    \item DDPG with dense rewards
\end{itemize}
Detailed hyperparameters can be found in~\Cref{appendix:hyper}.

\begin{figure}
    \centering
    \begin{subfigure}[b]{0.49\textwidth}
        \includegraphics[width=\textwidth]{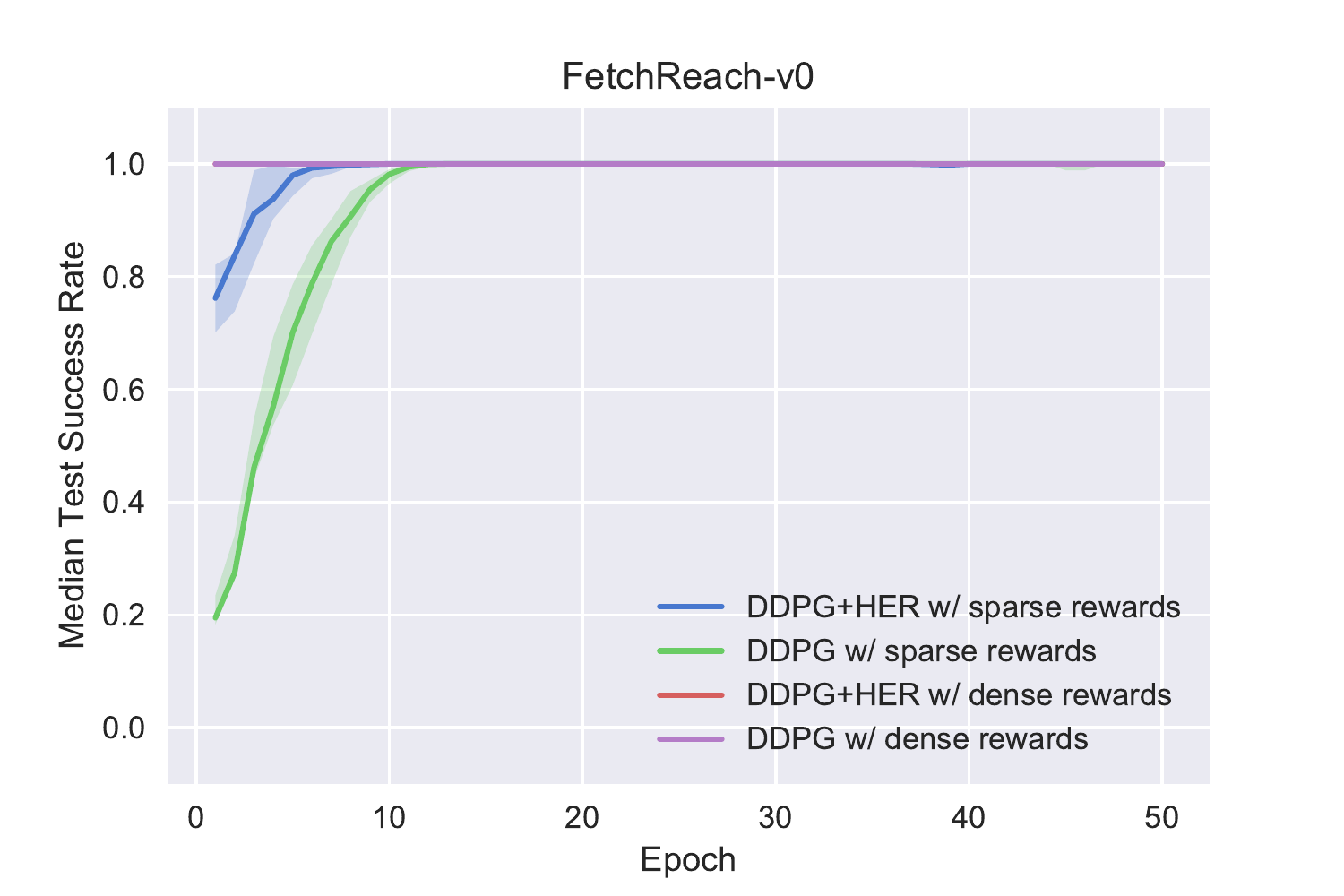}
    \end{subfigure}
    ~
    \begin{subfigure}[b]{0.49\textwidth}
        \includegraphics[width=\textwidth]{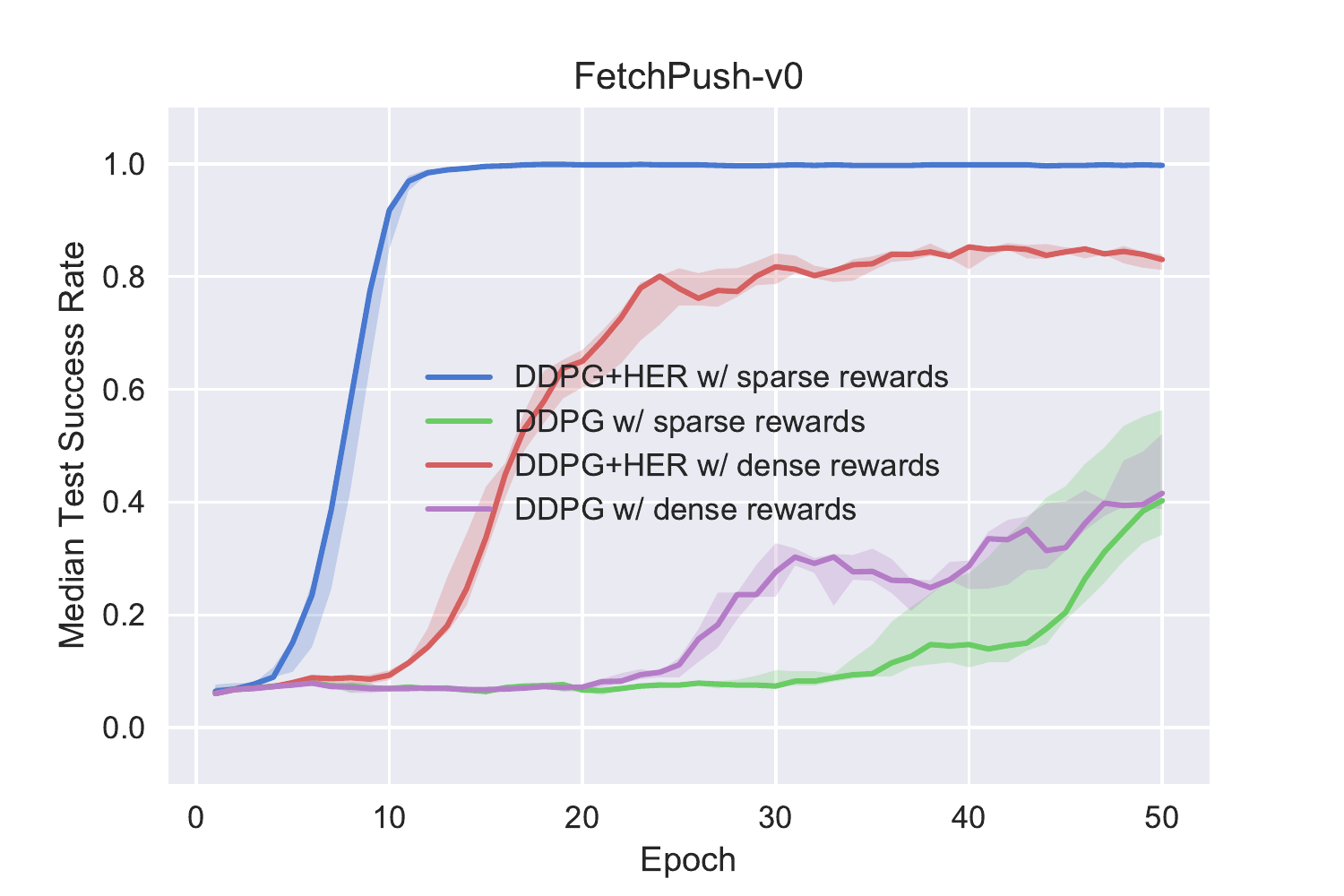}
    \end{subfigure}
    \\
    \begin{subfigure}[b]{0.49\textwidth}
        \includegraphics[width=\textwidth]{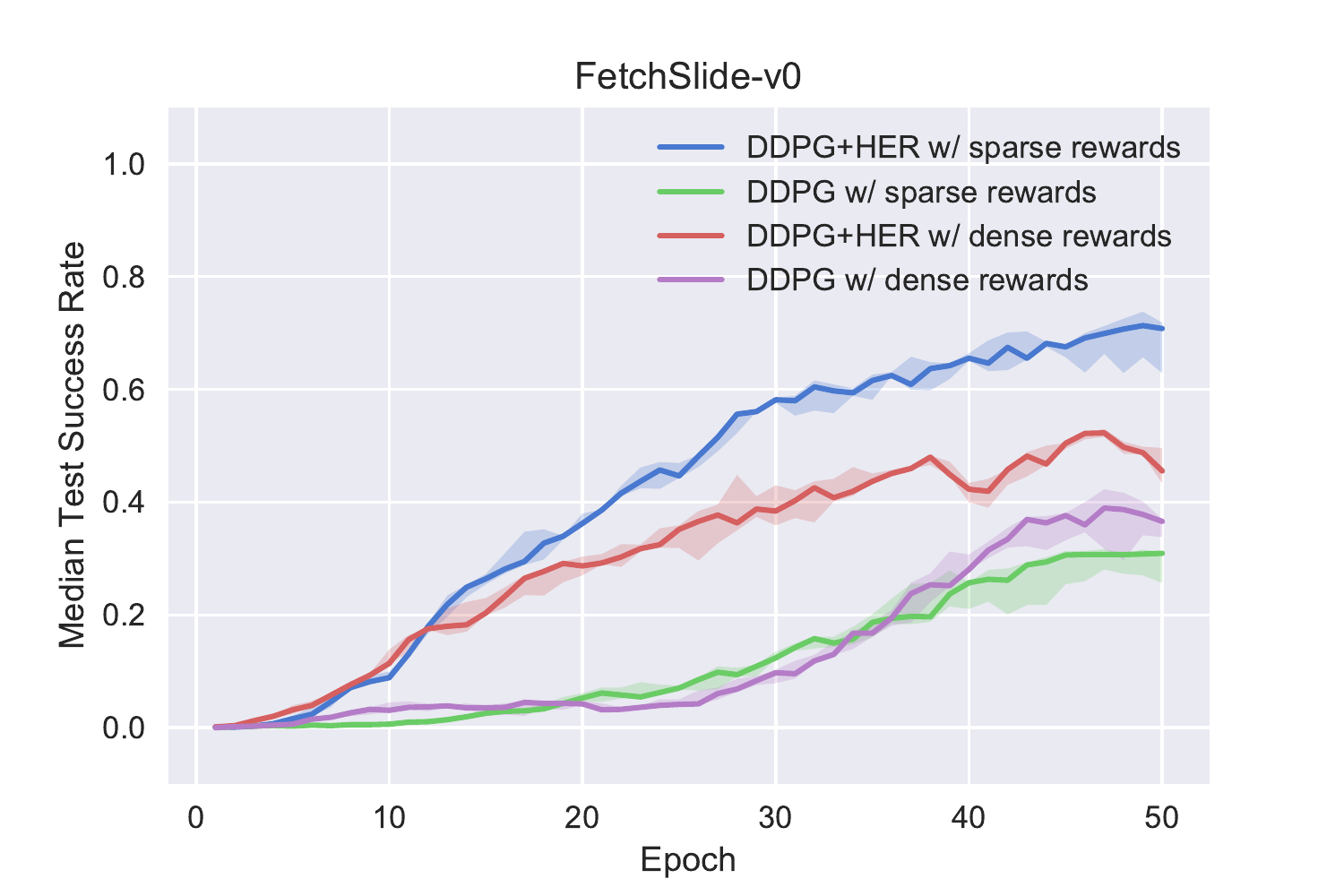}
    \end{subfigure}
    ~
    \begin{subfigure}[b]{0.49\textwidth}
        \includegraphics[width=\textwidth]{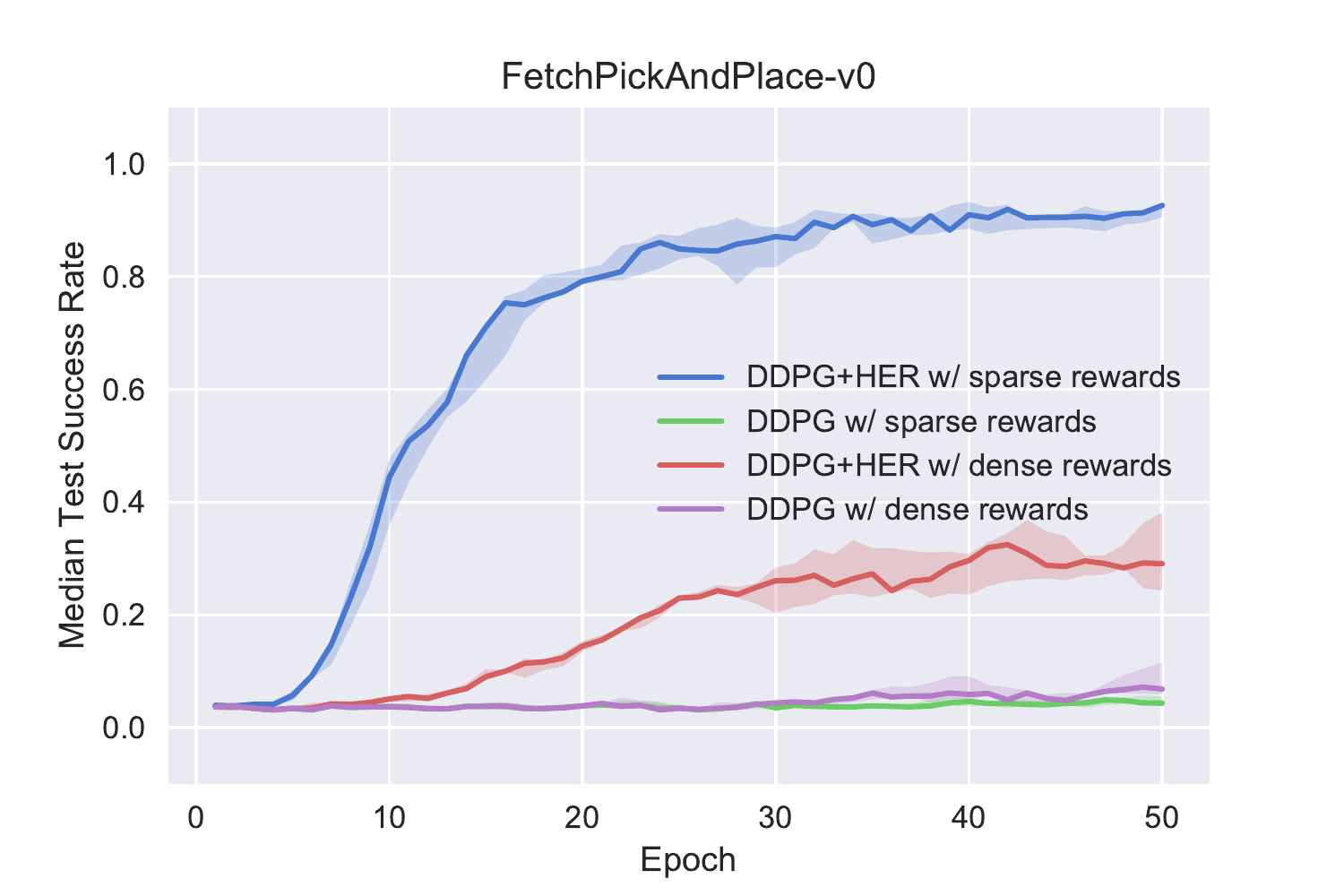}
    \end{subfigure}
    \caption{Median test success rate (line) with interquartile range (shaded area) for all four Fetch environments.}
    \label{fig:fetch-results}
\end{figure}

For all environments, we train on a single machine with $19$ CPU cores.
Each core generates experience using two parallel rollouts and uses MPI for synchronization.
For \texttt{FetchReach}, \texttt{FetchPush}, \texttt{FetchSlide}, \texttt{FetchPickAndPlace}, and \texttt{HandReach}, we train for $50$ epochs (one epoch consists of $19 \cdot 2 \cdot 50 = 1\,900$ full episodes), which amounts to a total of $4.75 \cdot 10^6$ timesteps.
For the remaining environments, we train for $200$ epochs, which amounts to a total of $38 \cdot 10^6$ timesteps.
We evaluate the performance after each epoch by performing $10$ deterministic test rollouts per MPI worker and then compute the test success rate by averaging across rollouts and MPI workers.
Our implementation is available as part of OpenAI Baselines\footnote{\url{https://github.com/openai/baselines}}~\citep{baselines}.
In all cases, we repeat an experiment with $5$ different random seeds and report results by computing the median test success rate as well as the interquartile range.

\Cref{fig:fetch-results} depicts the median test success rate for all four Fetch environments.
\texttt{FetchReach} is clearly a very simple environment and can easily be solved by all four configurations.
On the remaining environments, DDPG+HER clearly outperforms all other configurations.
Interestingly, DDPG+HER performs best if the reward structure is sparse but is also able to successfully learn from dense rewards.
For vanilla DDPG, it is typically easier to learn from dense rewards with sparse rewards being more challenging.

\begin{figure}
    \centering
    \begin{subfigure}[b]{0.49\textwidth}
        \includegraphics[width=\textwidth]{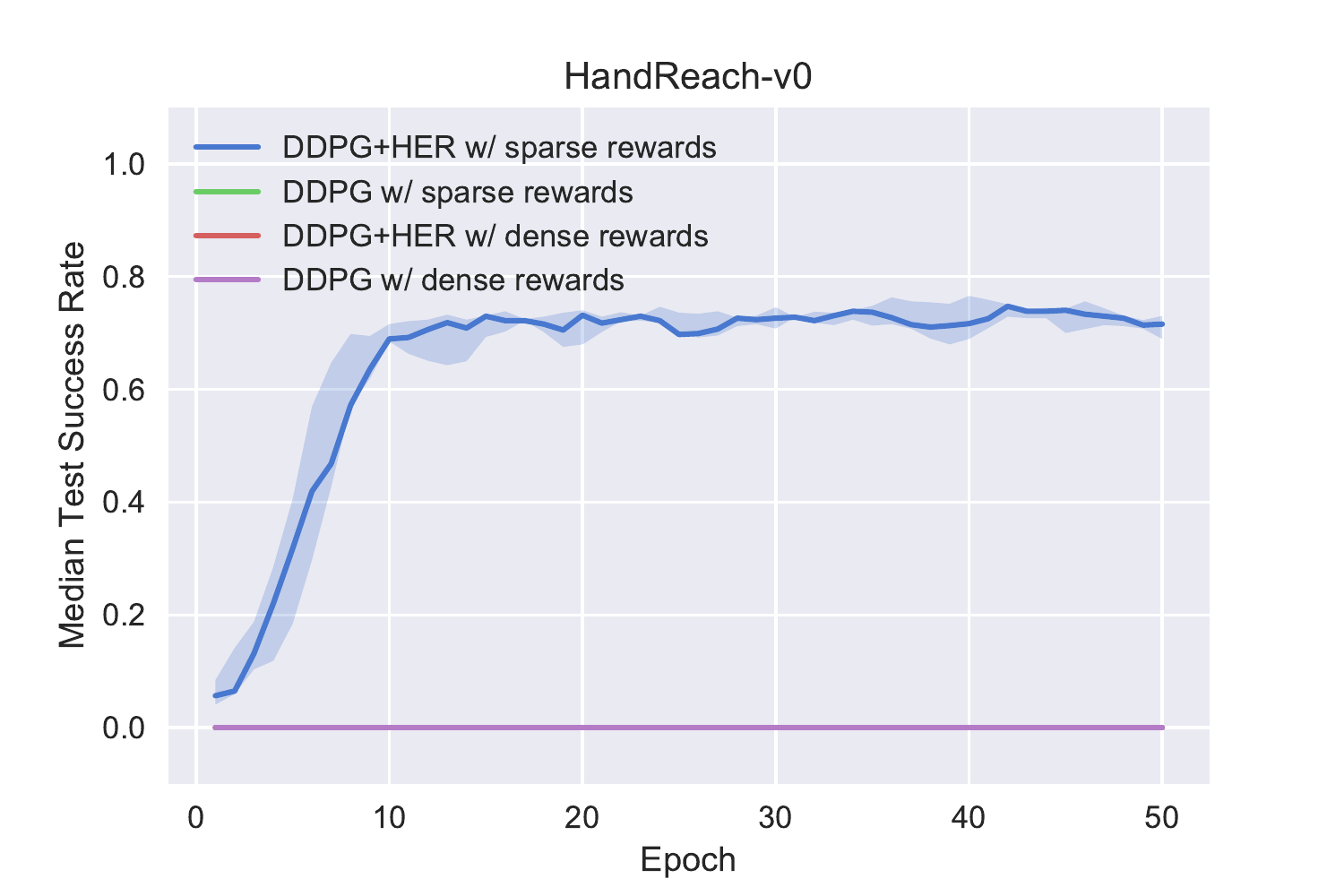}
    \end{subfigure}
    ~
    \begin{subfigure}[b]{0.49\textwidth}
        \includegraphics[width=\textwidth]{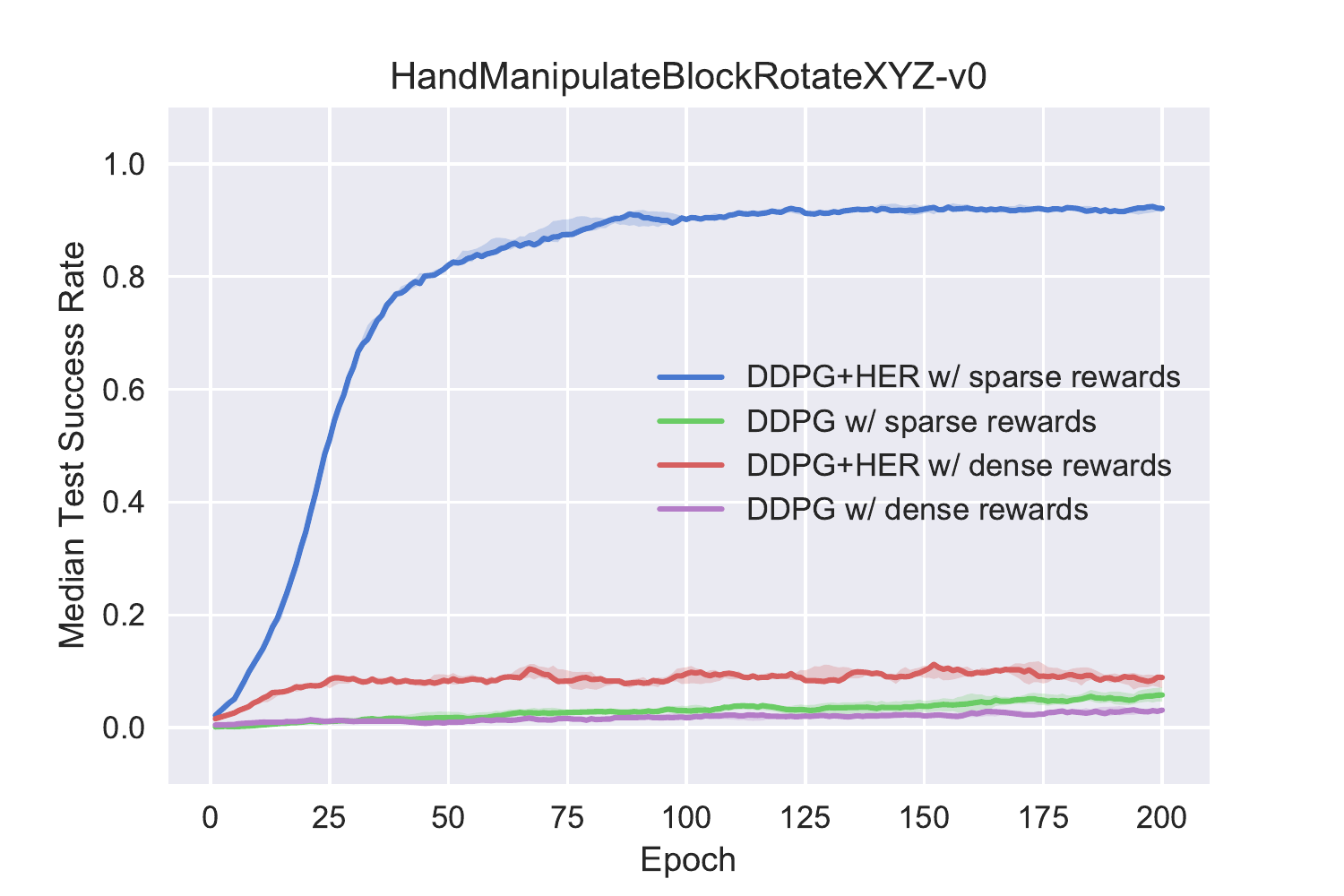}
    \end{subfigure}
    \\
    \begin{subfigure}[b]{0.49\textwidth}
        \includegraphics[width=\textwidth]{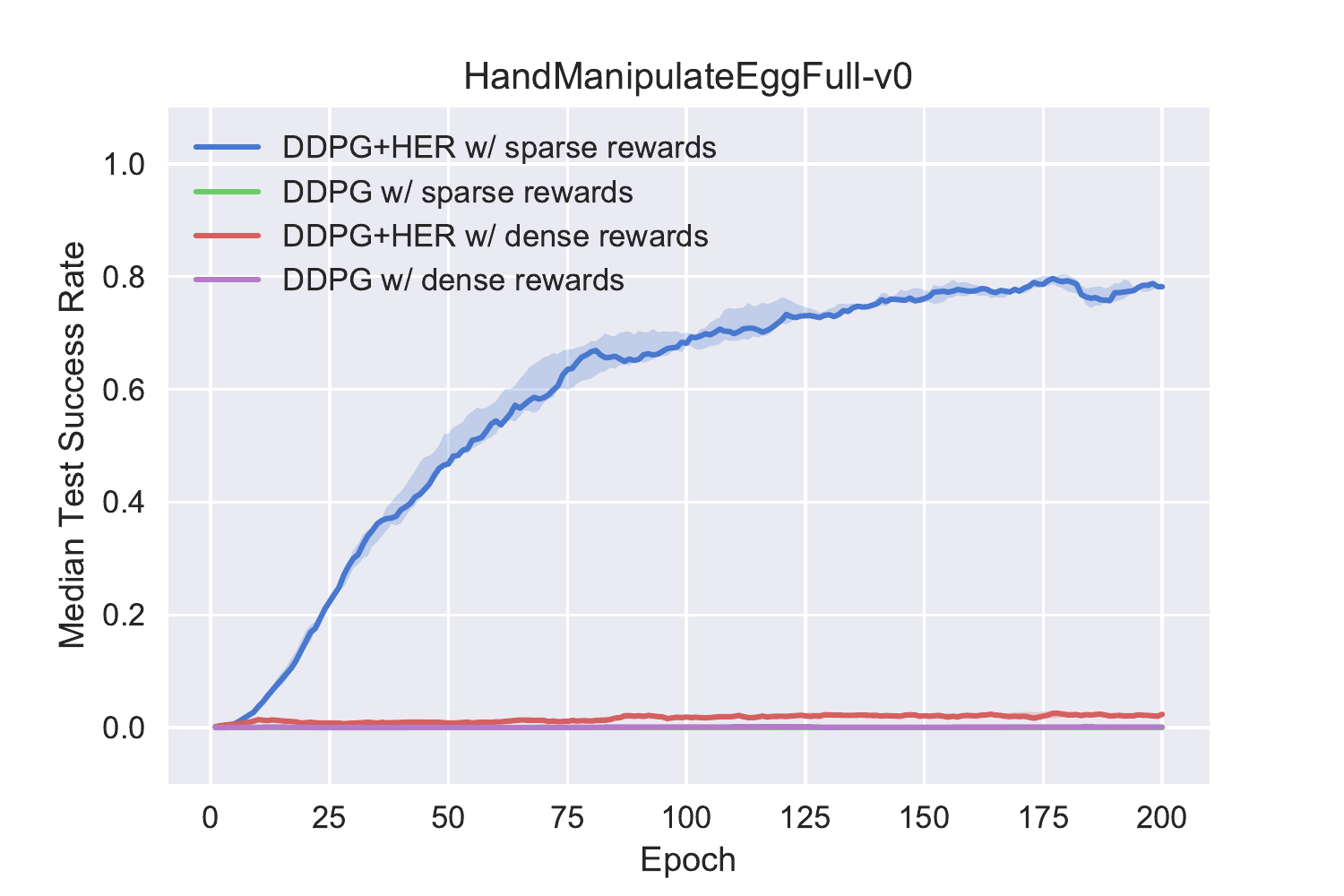}
    \end{subfigure}
    ~
    \begin{subfigure}[b]{0.49\textwidth}
        \includegraphics[width=\textwidth]{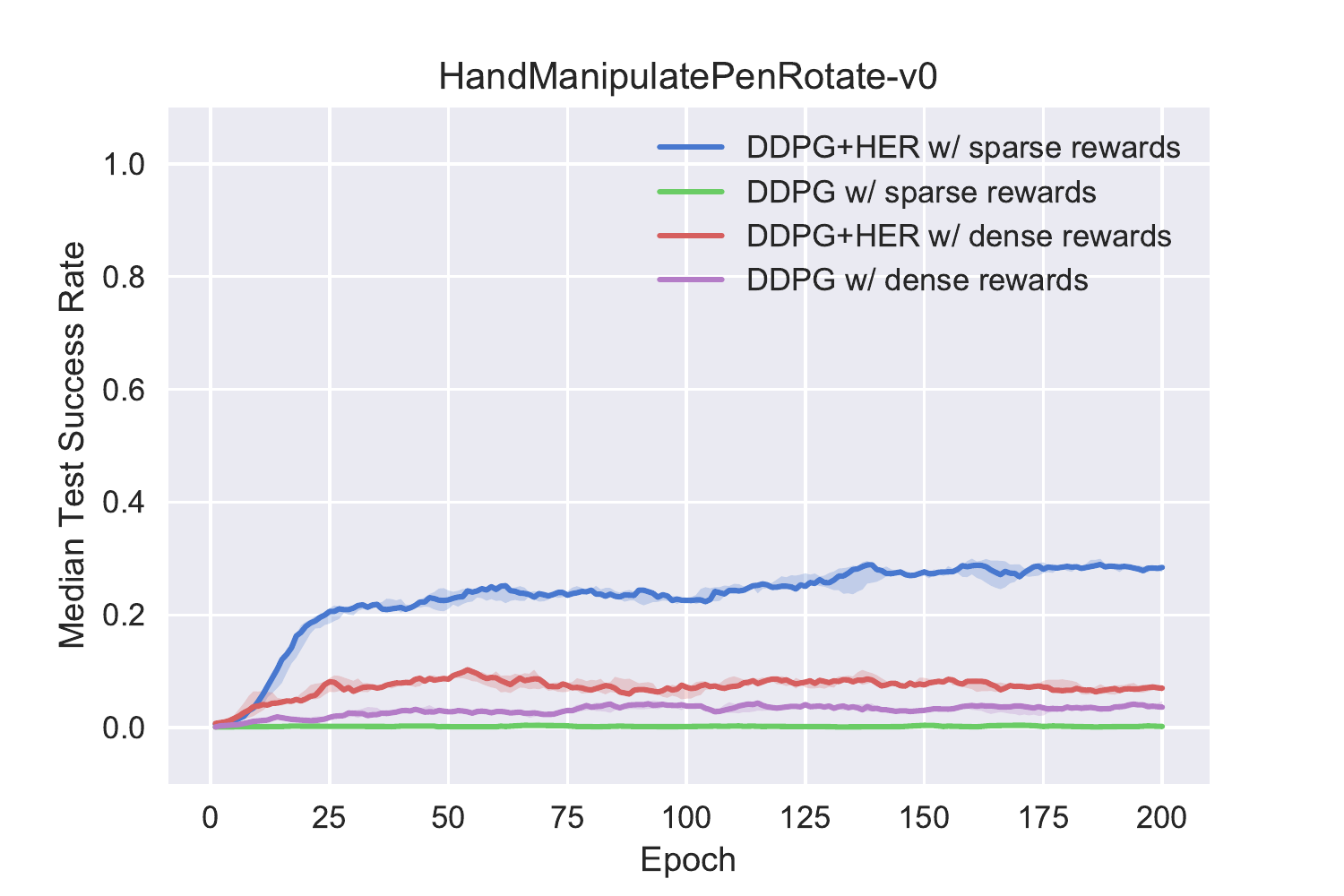}
    \end{subfigure}
    \caption{Median test success rate (line) with interquartile range (shaded area) for all four Fetch environments.}
    \label{fig:hand-results}
\end{figure}

\Cref{fig:hand-results} depicts the median test success rate for all four hand environments.
Similar to the Fetch environments, DDPG+HER significantly outperforms the DDPG baseline.
In fact, the baseline often is not able to learn the problem at all.
Similar to before, the sparse reward structure works significantly better than the dense reward when using HER.
HER is able to learn partly successful policies on all environments but especially \texttt{HandManipulatePen} is especially challenging and we are not able to fully solve it.
Note that we do not depict results for all variants of the four environments in this figure.
A complete set of plots for all environments and their variants can be found in \Cref{appendix:plots}.

We believe the reason why DDPG+HER typically performs better with sparse rewards is mainly due to the following two reasons:
\begin{itemize}
    \item Learning the critic is much simpler for sparse rewards. In the dense case, the critic has to approximate a highly non-linear function that includes the Euclidean distance between positions and the difference between two quaternions for rotations. On the other hand, learning the sparse return is much simpler since the critic only has to differentiate between successful and failed states.
    \item A dense reward biases the policy towards a specific strategy. For instance, it may be beneficial to first grasp an object properly and then start rotating it towards the desired goal. The dense reward however encourages the policy to chose a strategy that achieves the desired goal directly.
\end{itemize}
\section{Request for Research}

Deciding which problem is worth working on is probably the hardest part of doing research.
Below we present a set of research problems which we believe can lead to
widely-applicable RL improvements.
For each problem we propose at least one potential solution
but solving many of them will require inventing new ideas.
To make tracking the progress of work on these ideas easier,
we would like to ask authors to cite this report
when publishing related research.


\paragraph{Automatic hindsight goals generation}
In \cite{her} the goals used for HER were generated using a hand-crafted heuristic, e.g.
replaying with a goal which was achieved at a random future timestep in the episode.
Instead, we could learn which goals
are most valuable for replay. They could be chosen from the goals achieved or seen during training or generated by a separate
neural network given a transition as input.
The biggest question is how to judge which goals are most valuable for replay.
One option would be to train the generator to \emph{maximize} the Bellman error.
This bears a lot of similarity to Prioritized Experience Replay~\citep{schaul2015prioritized}
and we expect that some techniques from this paper may be useful here.

\paragraph{Unbiased HER}
HER changes the joint distribution of replayed $(\mbox{state},\,\mbox{action},\,\mbox{next\_state},\,\mbox{goal})$ tuples in an unprincipled way.
This could, in theory, make training impossible in extremely stochastic environment albeit we have not noticed this in practice.
Consider an environment in which there is a special action which takes the agent
to a random state and the episode ends after that.
Such an action would seem to be perfect in hindsight if we replay with the goal achieved by the agent in the future. How to avoid this problem? One potential approach would be to use importance sampling to cancel the sampling bias but this would probably lead to prohibitively high variance of the gradient.

\paragraph{HER+HRL}
Hierarchical Actor-Critic~\citep{hac} showed some promising results in applying HER in Hierarchical RL setup.
One possible extension of this work would be
to replace in hindsight not only goals, but also higher-level actions, e.g.
if the higher level asked the lower level to reach state A,
but some other state B was reached,
we could replay this episode replacing the higher-level action with B.
This could allow the higher level to learn even when the lower level policy is very bad but
is not very principled and could make training unstable.

\paragraph{Richer value functions}
UVFA~\citep{uvfa} extended value functions to multiple goals,
while TDM~\citep{tdm} extended them to different time horizons.
Both of these innovations can make training easier, despite the fact that the learned function is more complicated.
What else could we fed to the value function to improve the sample-efficiency?
How about discount factor or success threshold for binary rewards?

\paragraph{Faster information propagation}
Most state-of-the-art off-policy RL algorithms use target networks to stabilize training (e.g. DQN~\citep{dqn} or DDPG~\citep{ddpg}).
This, however, comes at a price of limiting the maximum learning speed of the algorithm as each target
network update sends the information about returns only one step backward in time (if one-step bootstrapping is used).
We noticed that the learning speed of DDPG+HER
in the early stages of training is often proportional
to the frequency of target network updates\footnote{Or to $1/(1-\mbox{averaging\_coefficient})$ if target networks is computed using a moving average of the main network's parameters.} but excessive frequency/magnitude of target network updates leads to unstable training and worse final performance.
How can we adapt the frequency of target network updates (or the moving average coefficient used to update the network)
to maximize the training speed?
Are there better ways to update the target network than a simple replacement or a moving average over time?
Are there other ways to stabilize training which does not limit the learning speed (e.g. clipped objective similar to the one used in PPO~\citep{ppo})?

\paragraph{HER + multi-step returns}
HER generates data which is extremely off-policy\footnote{Notice that normally off-policy data used by RL algorithms come
from an earlier version of the policy and therefores
is relatively close to on-policy data.
It is not the case for HER, because we completely replace
the goals which are fed to the network.}
and therefore multi-step returns can not be used
unless we employ some correction factors like importance sampling.
While there are many solutions for dealing with off-policies of the data (e.g. \cite{munos2016safe}),
it is not clear if they would perform well in the setup where the training data is so far from being on-policy.
Another approach would be to use multi-step optimality tightening inequalities~\citep{he2016learning}.
Using multi-step returns can be beneficial because the decreased frequency of bootstraping
can lead to less biased gradients. Moreover, it accelerates the
transfer of information about the returns backwards in time which, accordingly
to our experiment, is often the limiting factor in DDPG+HER training (compare previous paragraph).

\paragraph{On-policy HER}
How to combine HER with state-of-the-art on-policy RL algorithms like PPO~\citep{ppo}?
Some preliminary results with vanilla Policy Gradients were presented by \cite{rauber2017hindsight}, but this approach needs to be tested on more challenging environments like the ones proposed in this report.
One possible option would also be to to use techniques similar to the ones employed in IPG~\citep{ipg}.

\paragraph{Combine HER with recent improvements in RL}
It would be interesting to see how recent improvements in RL perform while combined with HER.
The list of potential improvements is long e.g. Prioritized Experience Replay~\citep{schaul2015prioritized},
distributional RL~\citep{bellemare2017distributional},
entropy-regularized RL~\citep{schulman2017equivalence},
or reverse curriculum generation~\citep{florensa2017reverse}.

\paragraph{RL with very frequent actions}
RL algorithms are very sensitive to the frequency of taking actions which is why
frame skip technique is usually used on Atari~\citep{dqn}.
In continuous control domains, the performance goes to zero as the frequency of taking actions goes to infinity,
which is caused by two factors: inconsistent exploration and the necessity to bootstrap more times
to propagate information about returns backward in time.
How to design a sample-efficient RL algorithm which can retain its performance even
when the frequency of taking actions goes to infinity?
The problem of exploration can be addressed by using parameters noise for exploration~\citep{plappert2017parameter}
and faster information propagation could be achieved by employing multi-step returns.
Other approach could be an adaptive and learnable frame skip.

\medskip
{
\small
\bibliography{paper}
}

\clearpage
\appendix
\section{Goal-based API Examples}
\label{appendix:code-examples}

\paragraph{Exposed reward function}
The following example demonstrates how the exposed reward function can be used to re-compute a reward with substituted goals.
The info dictionary can be used to store additional information that may be necessary to re-compute the reward but that is independent of the goal, e.g. state derived from the simulation.

\begin{lstlisting}[language=Python]
import gym
 
env = gym.make('FetchReach-v0')
env.reset()
obs, reward, done, info = env.step(
    env.action_space.sample())

# The following always has to hold:
assert reward == env.compute_reward(
    obs['achieved_goal'], obs['desired_goal'], info)

# ... but you can also substitute goals:
substitute_goal = obs['achieved_goal'].copy()
substitute_reward = env.compute_reward(
    obs['achieved_goal'], substitute_goal, info)
\end{lstlisting}

\paragraph{Compatibility with standard RL algorithms}
The following example demonstrates how to wrap the new goal-based environments to make their observation spaces compatible with existing implementations.
To do so, simply wrap any goal-based environment with \texttt{gym.wrappers.FlattenDictWrapper} and specify the desired keys of the dictionary that you would like to use.

\begin{lstlisting}[language=Python]
import gym
 
env = gym.make('FetchReach-v0')
print(type(env.reset())) 
# prints "<class 'dict'>"

env = gym.wrappers.FlattenDictWrapper(
    env, ['observation', 'desired_goal'])
ob = env.reset()
print(type(ob), ob.shape)
# prints "<class 'numpy.ndarray'> (13,)"
\end{lstlisting}

\section{Hyperparameters}
\label{appendix:hyper}
To ensure a fair comparison, we perform a hyperparameter search over the following parameters:
\begin{itemize}
    \item Actor learning rate: $\{1\cdot10^{-4}, 3\cdot10^{-4}, 6\cdot10^{-4}, 1\cdot10^{-3}, 3\cdot10^{-3}, 6\cdot10^{-3}, 1\cdot10^{-2}\}$
    \item Critic learning rate: $\{1\cdot10^{-4}, 3\cdot10^{-4}, 6\cdot10^{-4}, 1\cdot10^{-3}, 3\cdot10^{-3}, 6\cdot10^{-3}, 1\cdot10^{-2}\}$
    \item Polyak-averaging coefficient $\tau$: $\{0.9, 0.93, 0.95, 0.97, 0.99\}$
    \item Batch size: $\{32, 64, 128, 256\}$
    \item Probability of random action: $\{0, 0.1, 0.2, 0.3, 0.4\}$
    \item Scale of additive Gaussian noise: $\sigma$: $\{0, 0.1, 0.2, 0.3, 0.4\}$
    \item Action L2 norm coefficient: $\{0, 0.01, 0.03, 0.1, 0.3, 0.6, 1.\}$
\end{itemize}
Since searching all possible combinations exhaustivley is intractable, we randomly sample $40$~combinations and train a policy on the \texttt{HandManipulateBlockRotateZ-v0} environment for all four configurations (DDPG+HER sparse, DDPG+HER dense, DDPG sparse, DDPG dense).
We picked this environment since all configurations are capable of learning on this environment.
For each configuration and combination we train with $3$ random seeds and average performance across this.
To select the best hyperparameter combination, we numerically compute the area under the test success rate curve and select the combination that achieves the best performance across all tasks.

All experiments in this paper use the following hyperparameters, which have been found by the aforementioned search:
\begin{itemize}
    \item Actor and critic networks: $3$ layers with $256$ units each and ReLU non-linearities
    \item Adam optimizer~\citep{adam} with $1\cdot10^{-3}$ for training both actor and critic
    \item Buffer size: $10^6$ transitions
    \item Polyak-averaging coefficient: $0.95$
    \item Action L2 norm coefficient: $1.0$
    \item Observation clipping: $[-200, 200]$
    \item Batch size: $256$
    \item Rollouts per MPI worker: $2$
    \item Number of MPI workers: $19$
    \item Cycles per epoch: $50$
    \item Batches per cycle: $40$
    \item Test rollouts per epoch: $10$
    \item Probability of random actions: $0.3$
    \item Scale of additive Gaussian noise: $0.2$
    \item Probability of HER experience replay: $0.8$
    \item Normalized clipping: $[-5, 5]$
\end{itemize}
All hyperparameters are described in greater detail in \cite{her}.

\section{Full Benchmark Results}
\label{appendix:plots}

\begin{figure}[h]
    \centering
    \includegraphics[width=\textwidth]{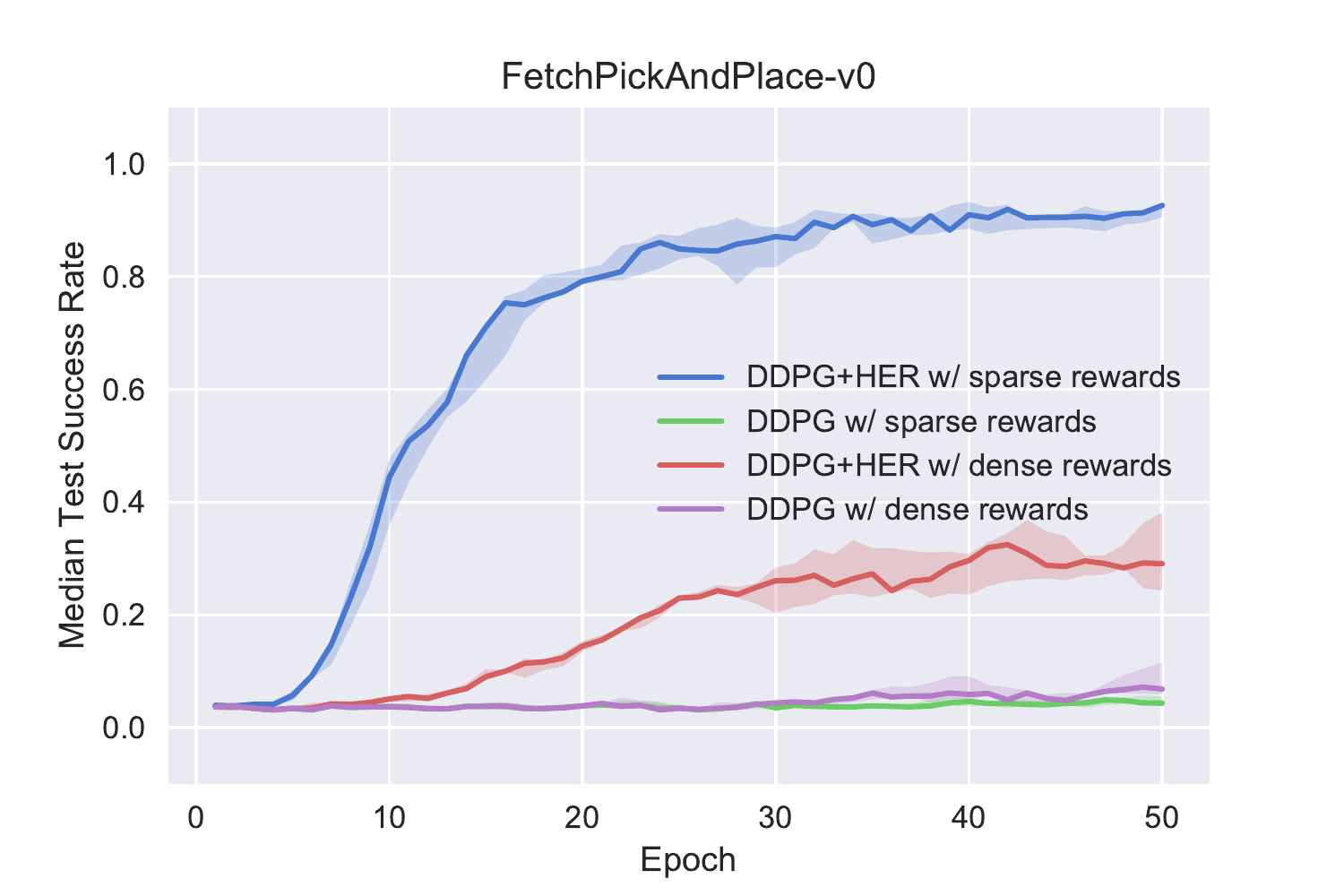}
\end{figure}

\begin{figure}[h]
    \centering
    \includegraphics[width=\textwidth]{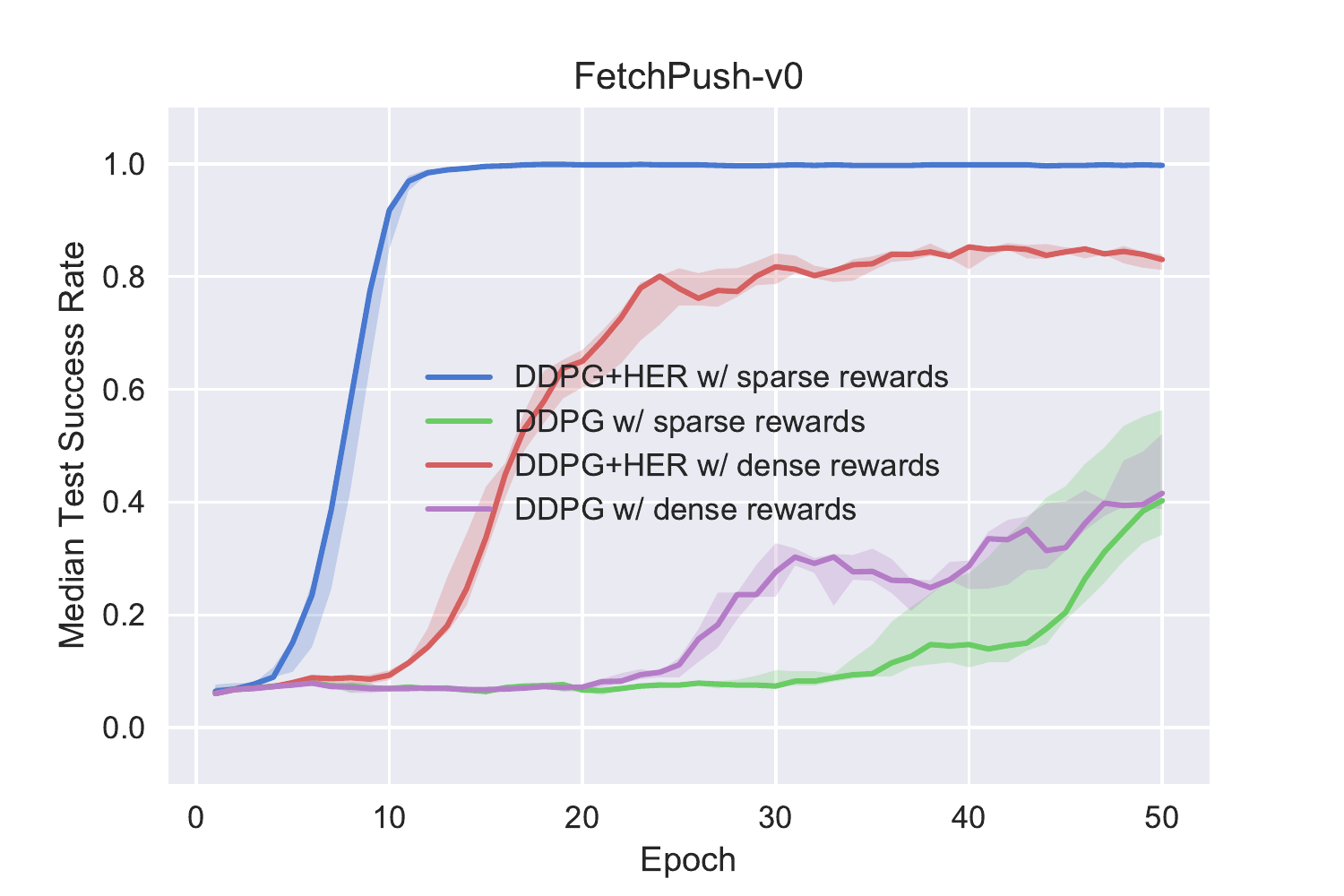}
\end{figure}

\begin{figure}[h]
    \centering
    \includegraphics[width=\textwidth]{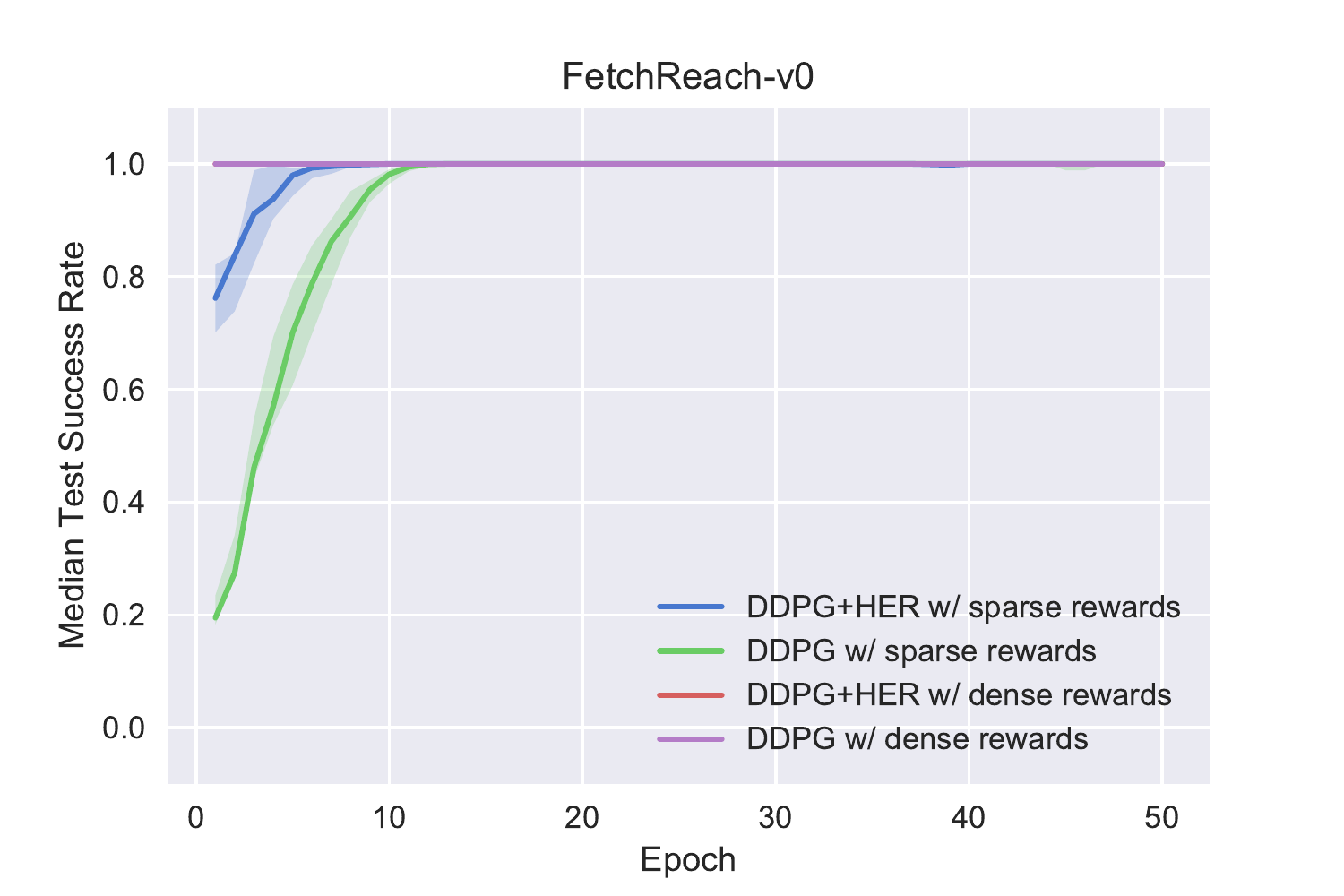}
\end{figure}

\begin{figure}[h]
    \centering
    \includegraphics[width=\textwidth]{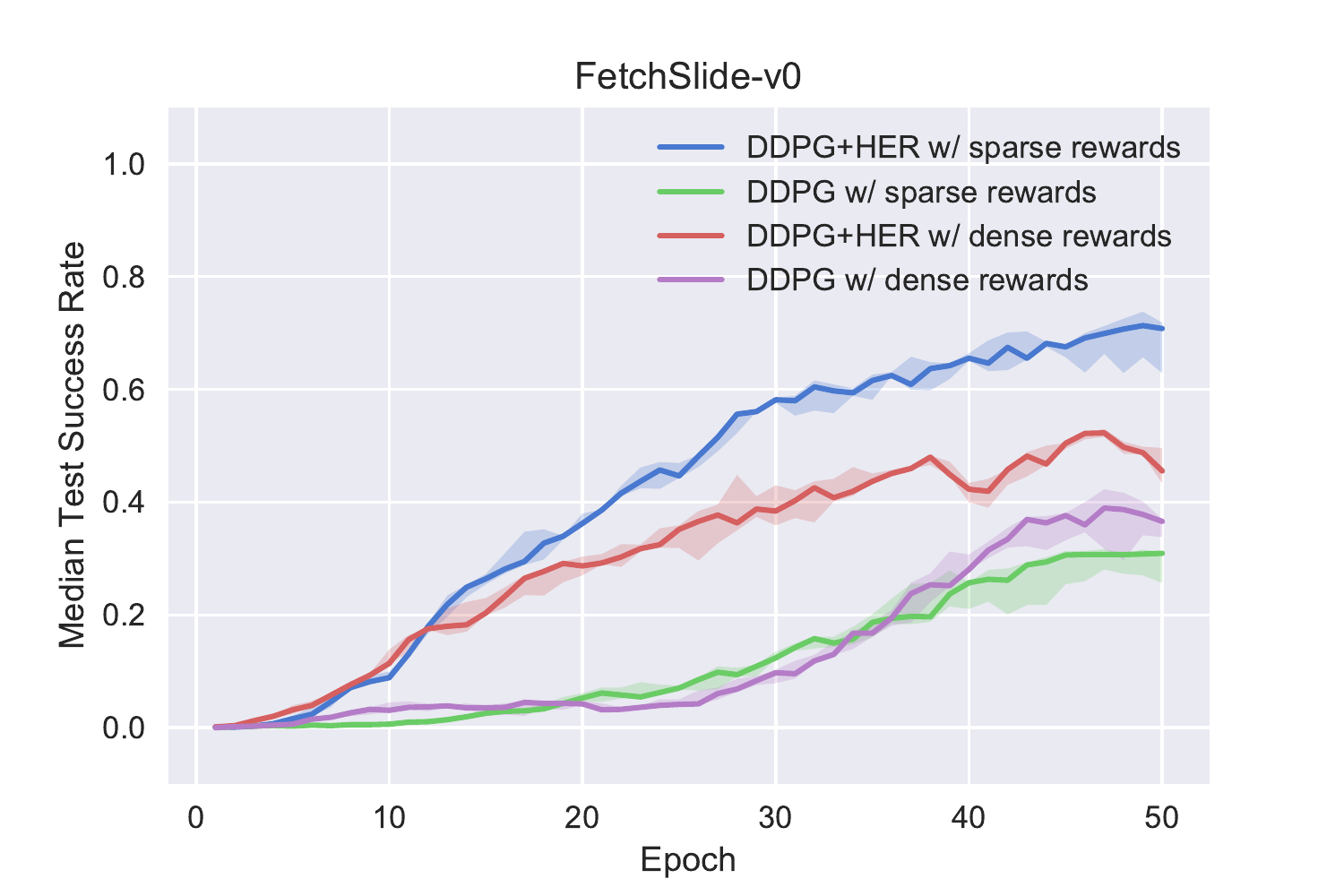}
\end{figure}

\begin{figure}[h]
    \centering
    \includegraphics[width=\textwidth]{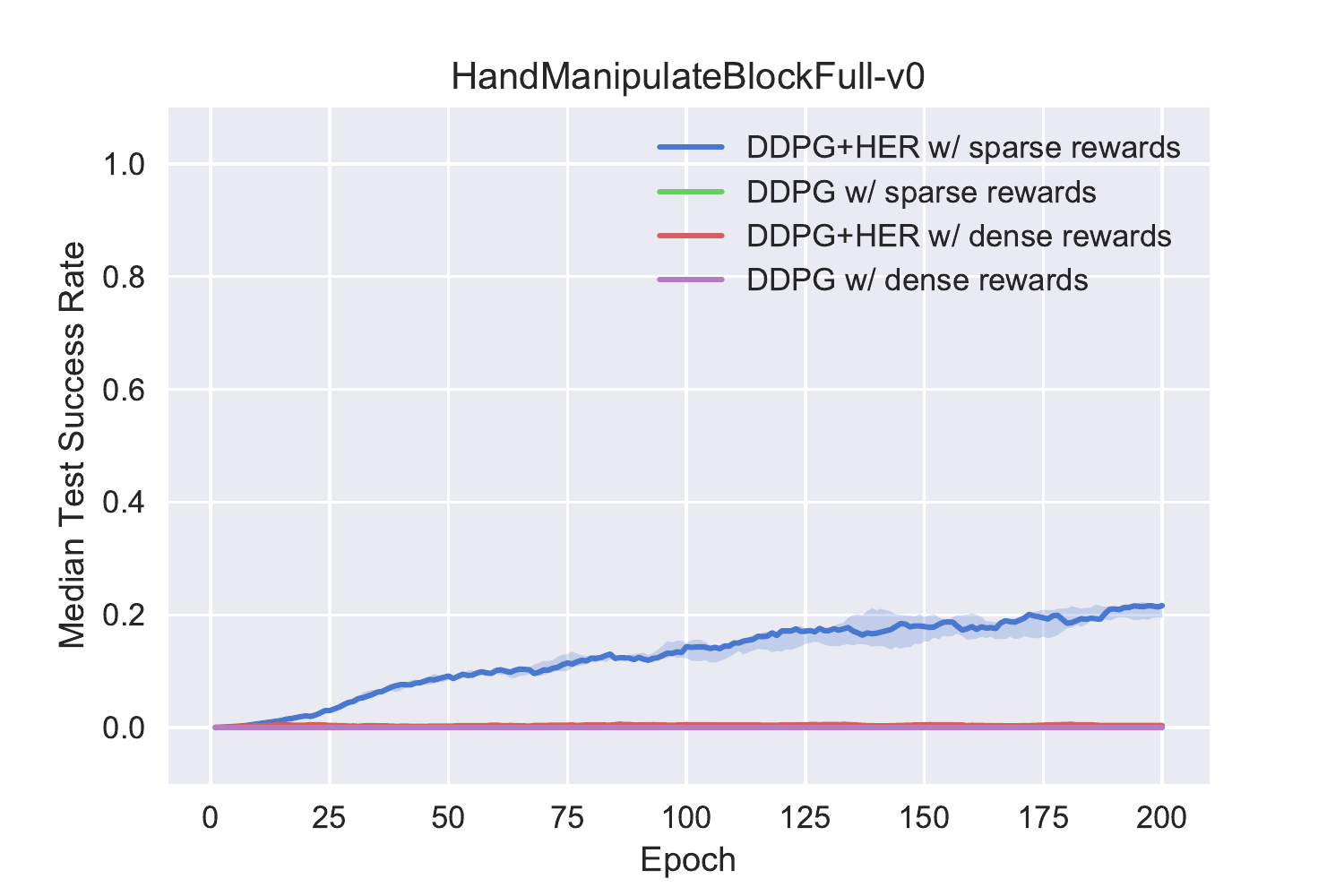}
\end{figure}

\begin{figure}[h]
    \centering
    \includegraphics[width=\textwidth]{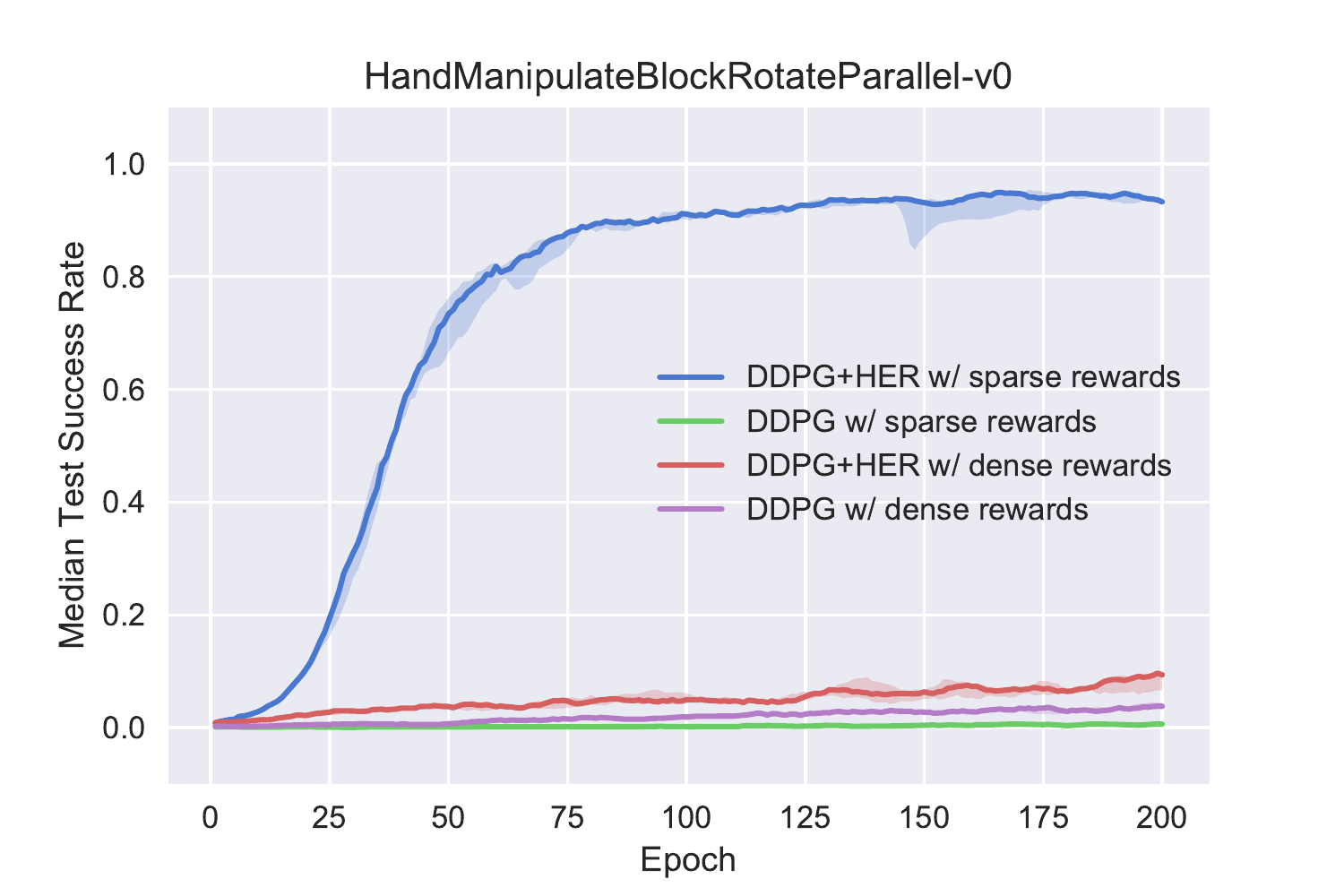}
\end{figure}

\begin{figure}[h]
    \centering
    \includegraphics[width=\textwidth]{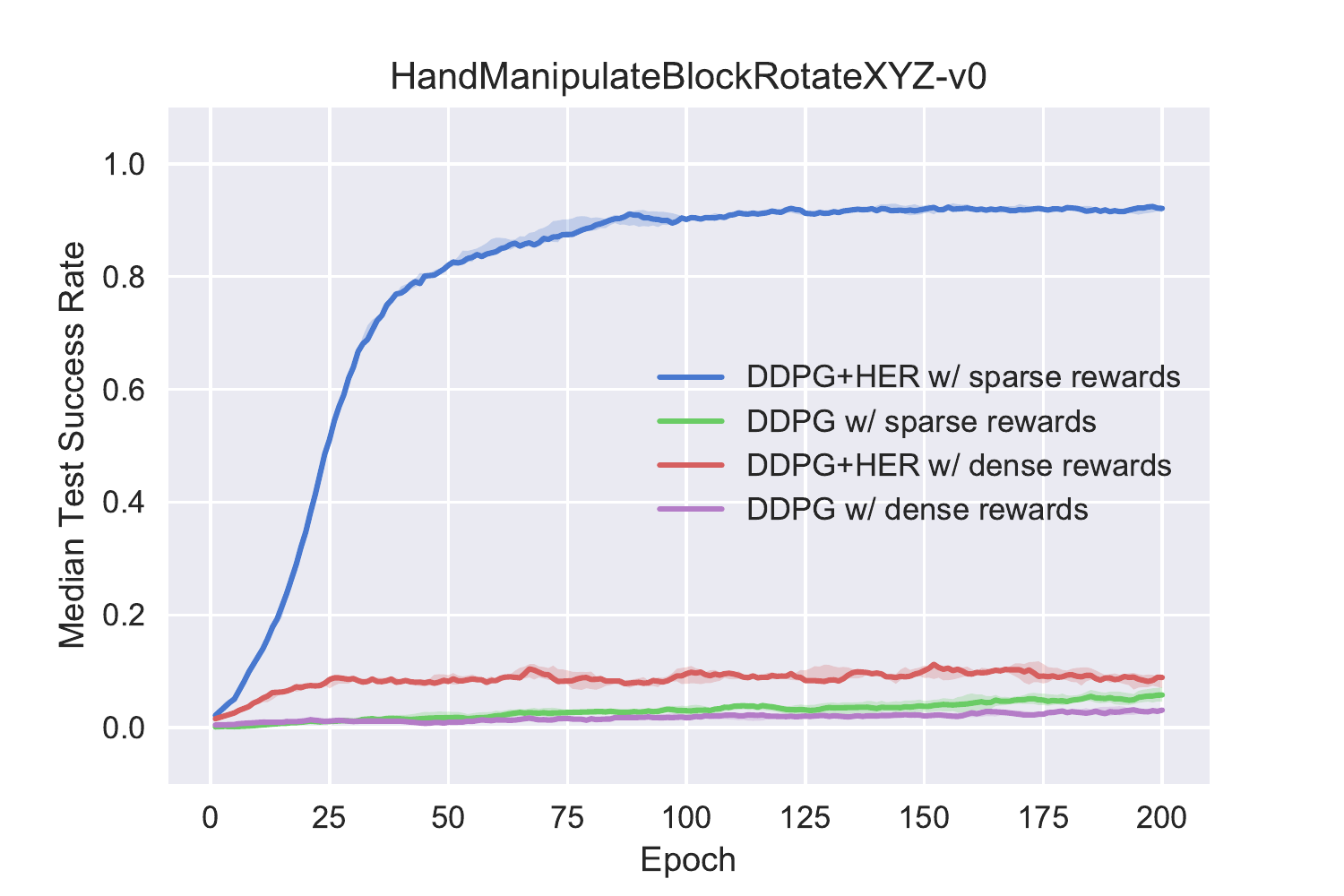}
\end{figure}

\begin{figure}[h]
    \centering
    \includegraphics[width=\textwidth]{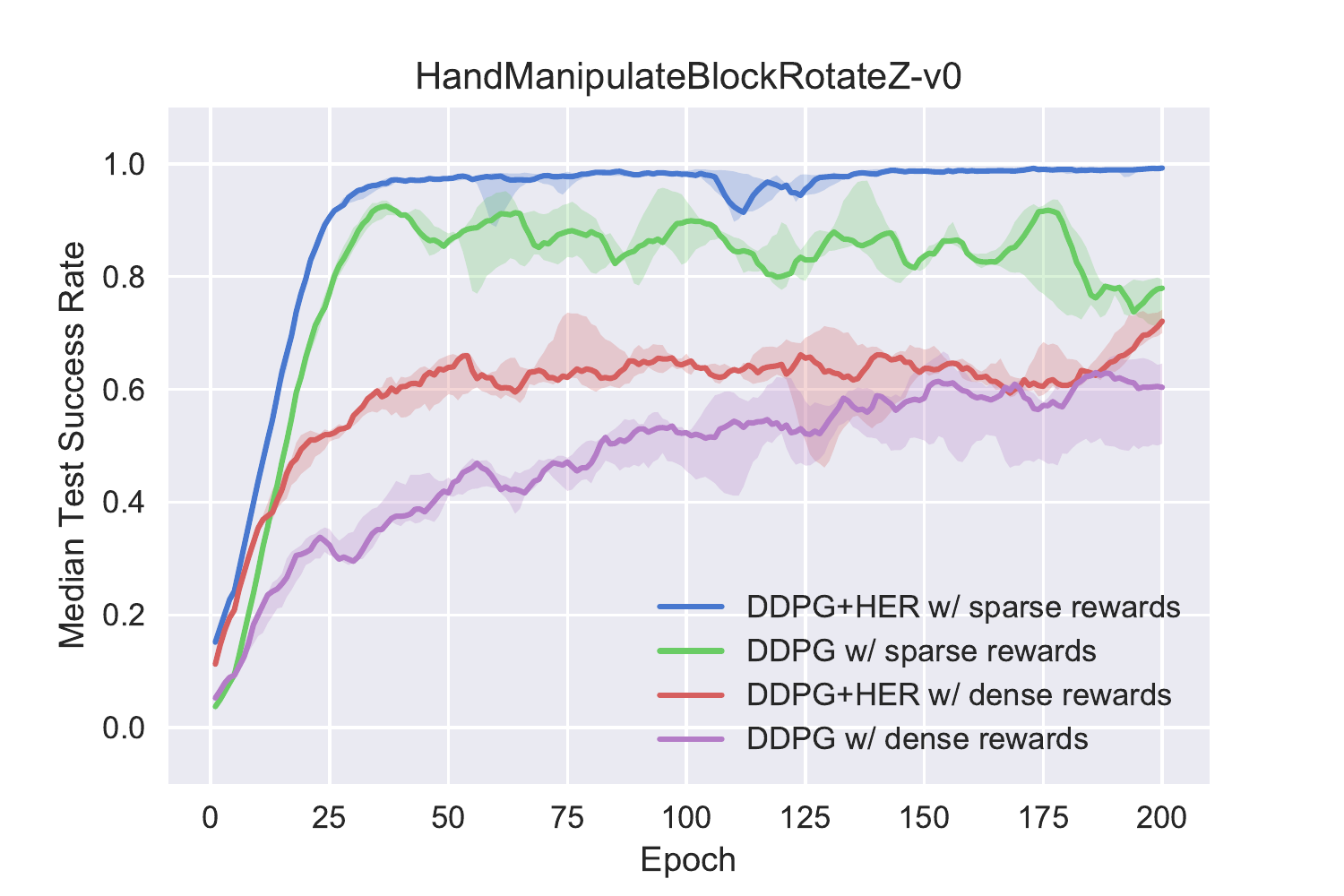}
\end{figure}

\begin{figure}[h]
    \centering
    \includegraphics[width=\textwidth]{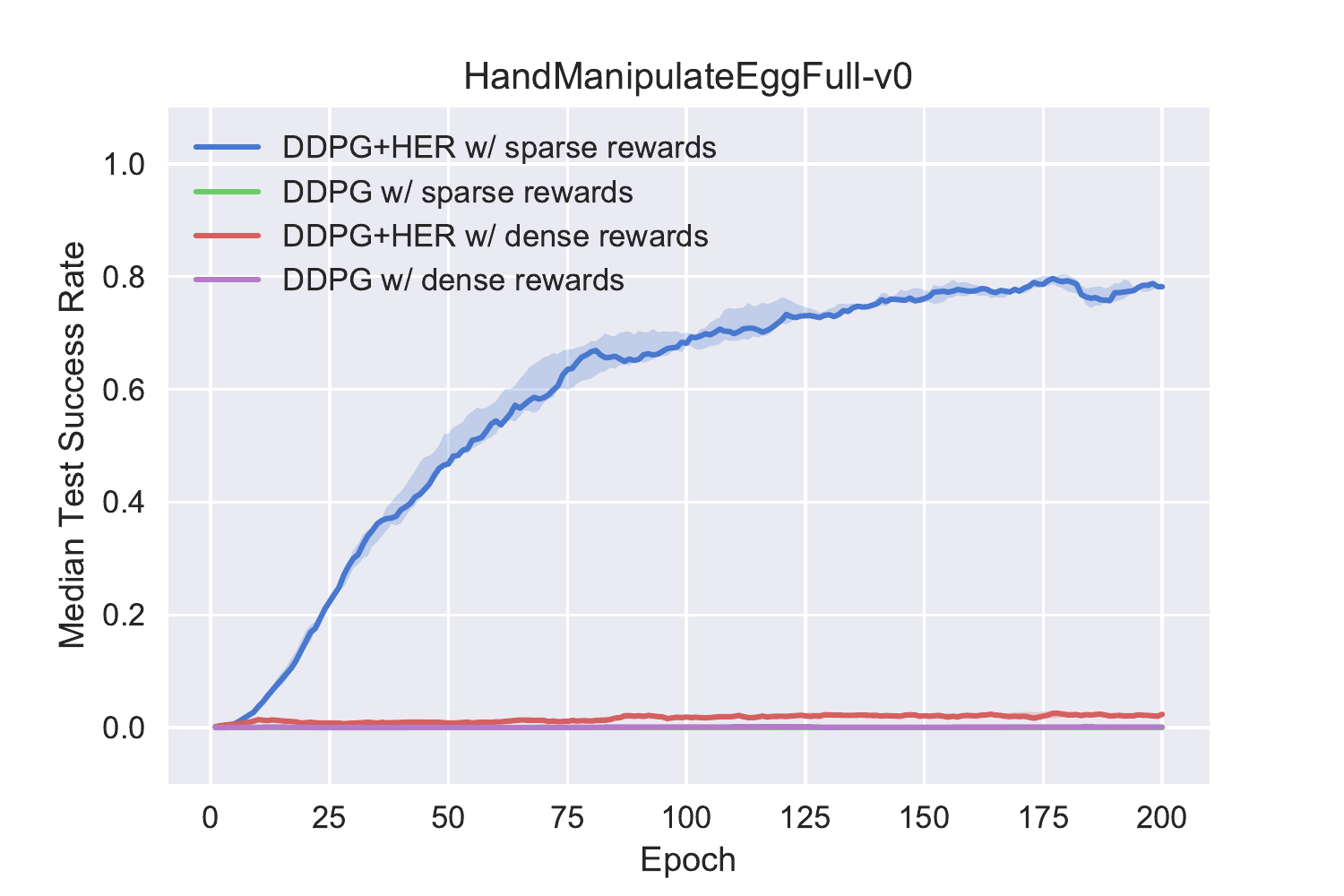}
\end{figure}

\begin{figure}[h]
    \centering
    \includegraphics[width=\textwidth]{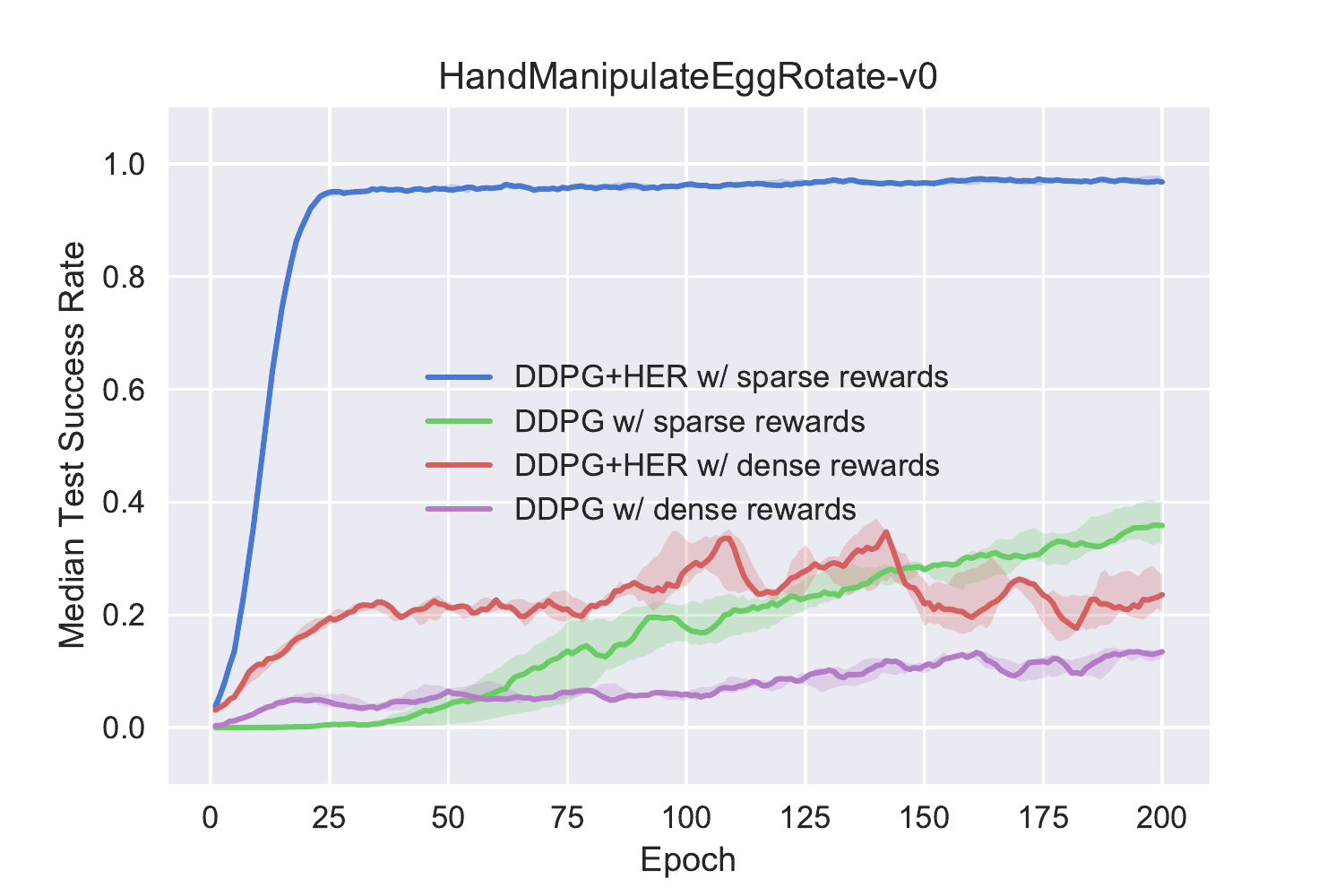}
\end{figure}

\begin{figure}[h]
    \centering
    \includegraphics[width=\textwidth]{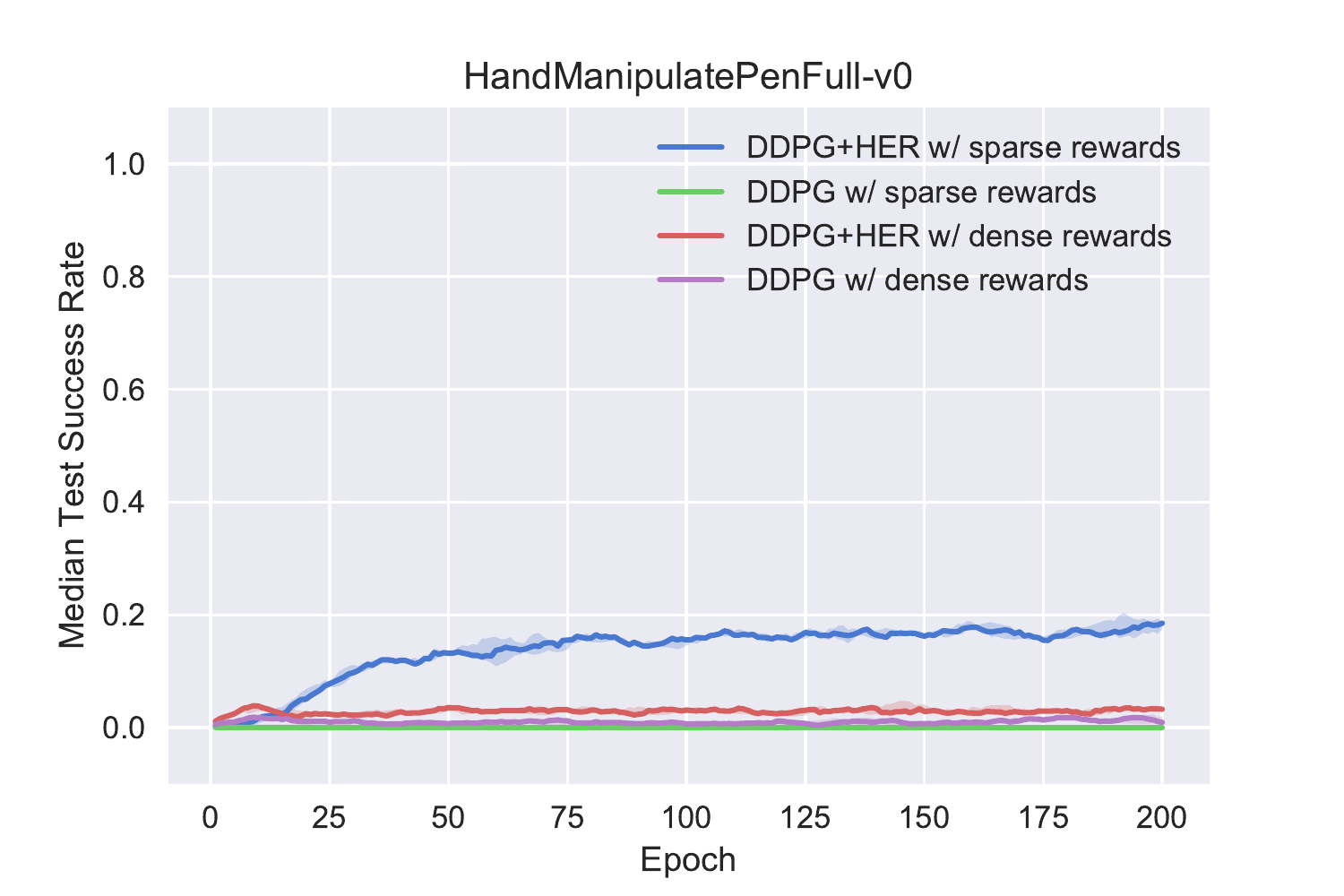}
\end{figure}

\begin{figure}[h]
    \centering
    \includegraphics[width=\textwidth]{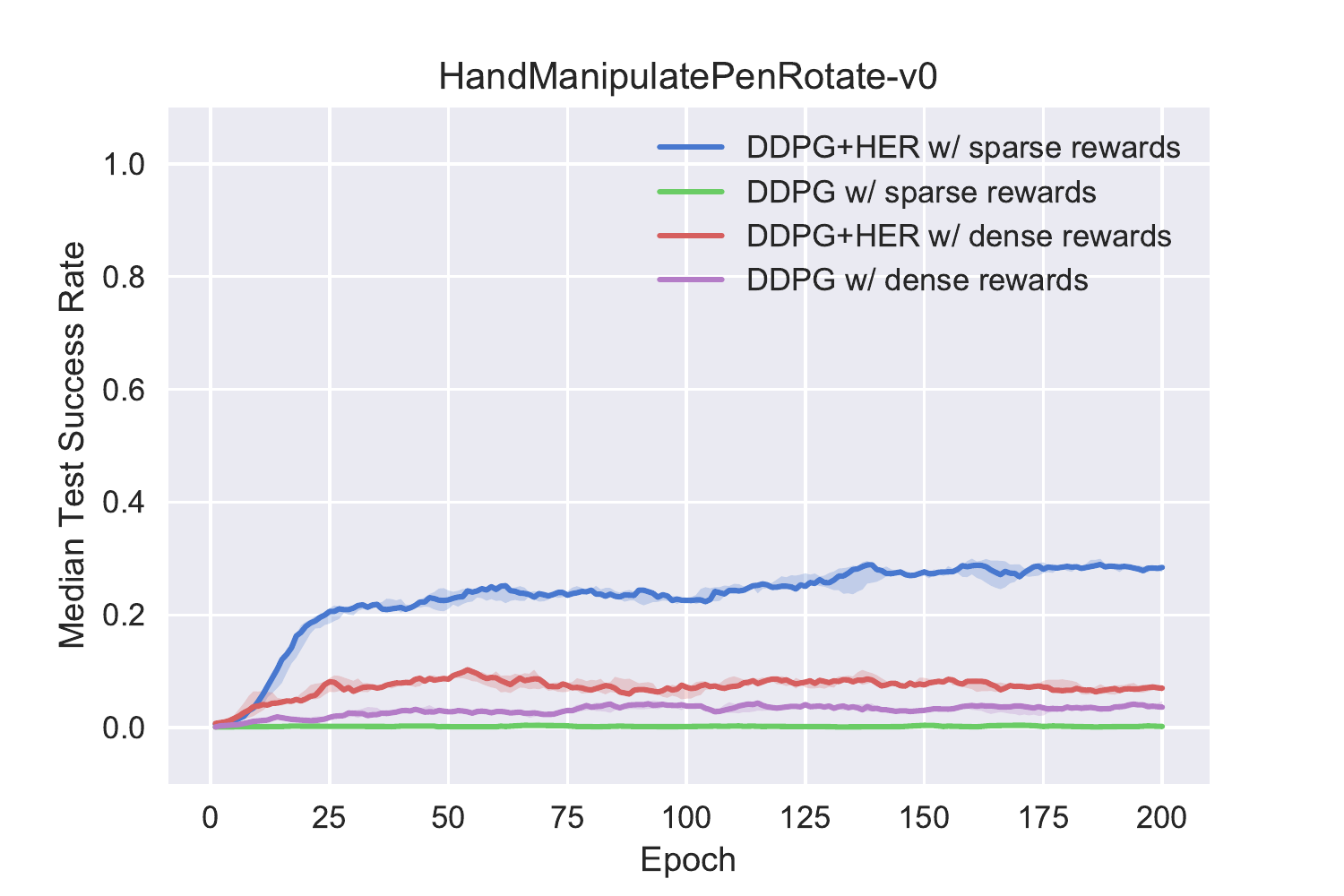}
\end{figure}

\end{document}